%% file: neurips_2025.tex
\documentclass{article}

\usepackage[preprint]{neurips_2025}

\usepackage[utf8]{inputenc} 
\usepackage[T1]{fontenc}    
\usepackage{hyperref}       
\usepackage{url}            
\usepackage{booktabs}       
\usepackage{amsfonts}       
\usepackage{nicefrac}       
\usepackage{microtype}      
\usepackage{xcolor}         

\usepackage{url}
\usepackage{graphicx}
\usepackage{xspace}
\usepackage{colortbl}
\usepackage{booktabs}
\usepackage{adjustbox}
\usepackage{multirow}
\usepackage{marvosym}
\usepackage{amsmath}
\usepackage{wrapfig}
\usepackage{cleveref}
\usepackage{natbib}

\usepackage{xcolor}

\definecolor{ForestGreen}{rgb}{0, 0.69, 0.31}
\definecolor{NavyBlue}{rgb}{0, 0.44, 0.75}
\newcommand{\hgreen}[1]{\textcolor{ForestGreen}{\textbf{#1}}} 
\newcommand{\hblue}[1]{\textcolor{NavyBlue}{\textbf{#1}}} 

\usepackage{pifont}

\newcommand{\xmark}{\ding{55}}
\newcommand{\methodname}{Visual-ARFT\xspace}
\newcommand{\benchname}{MAT\xspace}
\newcommand{\codename}{MAT-Coding\xspace}
\newcommand{\searchname}{MAT-Search\xspace}
\definecolor{Cerulean}{rgb}{0.0, 0.48, 0.65}

\title{Visual Agentic Reinforcement Fine-Tuning}

\author{
Ziyu Liu$^{1,2}$ \quad
Yuhang Zang$^{2}$\textsuperscript{\Letter}  \quad  
Yushan Zou$^{4}$ \quad
Zijian Liang$^{1}$ \quad \\
\textbf{Xiaoyi Dong$^{2,3}$} \quad 
\textbf{Yuhang Cao$^{2}$}  \quad
\textbf{Haodong Duan$^{2}$} \quad
\textbf{Dahua Lin$^{2,3}$}  \quad
\textbf{Jiaqi Wang$^{2}$\textsuperscript{\Letter}} \\
$^{1}$Shanghai Jiaotong University \quad $^{2}$Shanghai Artificial Intelligence Laboratory \\
$^3$The Chinese University of Hong Kong \quad $^4$Wuhan University \quad \\
{\tt\small liuziyu77@sjtu.edu.cn, \{zangyuhang, wangjiaqi\}@pjlab.org.cn}\\
{\tt\small \url{https://github.com/Liuziyu77/Visual-RFT/tree/main/Visual-ARFT}}  
\vspace{-2mm}
}

\begin{document}

\maketitle

\input{sec/0_abstract}

\input{sec/1_introduction}
\input{sec/2_related_work}
\input{sec/3_methodology}
\input{sec/4_benchmark}
\input{sec/4_experiments}
\input{sec/5_conclusion}


\newpage
{
    \small
    \bibliographystyle{plain}
    \bibliography{main}
}


\newpage

\appendix

\section*{\centering Appendix of Visual Agentic Reinforcement Fine-Tuning}
\input{sec/6_appendix.tex}



\end{document}

%% file: sec/0_abstract.tex
\begin{abstract}
A key trend in Large Reasoning Models (e.g., OpenAI's o3) is the native agentic ability to use external tools such as web browsers for searching and writing/executing code for image manipulation to think with images.
In the open-source research community, while significant progress has been made in language-only agentic abilities such as function calling and tool integration, the development of multi-modal agentic capabilities that involve truly thinking with images, and their corresponding benchmarks, are still less explored.
This work highlights the effectiveness of \textbf{V}isual \textbf{A}gentic \textbf{R}einforcement \textbf{F}ine-\textbf{T}uning (\textbf{Visual-ARFT}) for enabling flexible and adaptive reasoning abilities for Large Vision-Language Models (LVLMs).
With Visual-ARFT, open-source LVLMs gain the ability to browse websites for real-time information updates and write code to manipulate and analyze input images through cropping, rotation, and other image processing techniques.
We also present a \textbf{M}ulti-modal \textbf{A}gentic \textbf{T}ool Bench (\textbf{MAT}) with two settings (MAT-Search and MAT-Coding) designed to evaluate LVLMs' agentic search and coding abilities.
Our experimental results demonstrate that Visual-ARFT outperforms its baseline by +18.6\% F1 / +13.0\% EM on \codename and +10.3\% F1 / +8.7\% EM on \searchname, ultimately surpassing GPT-4o.
Visual-ARFT also achieves +29.3 F1\% / +25.9\% EM gains on existing multi-hop QA benchmarks such as 2Wiki and HotpotQA, demonstrating strong generalization capabilities.
Our findings suggest that Visual-ARFT offers a promising path toward building robust and generalizable multimodal agents.
\end{abstract}

%% file: sec/1_introduction.tex
\begin{figure}[h]
    \centering
   \includegraphics[width=.98\linewidth]{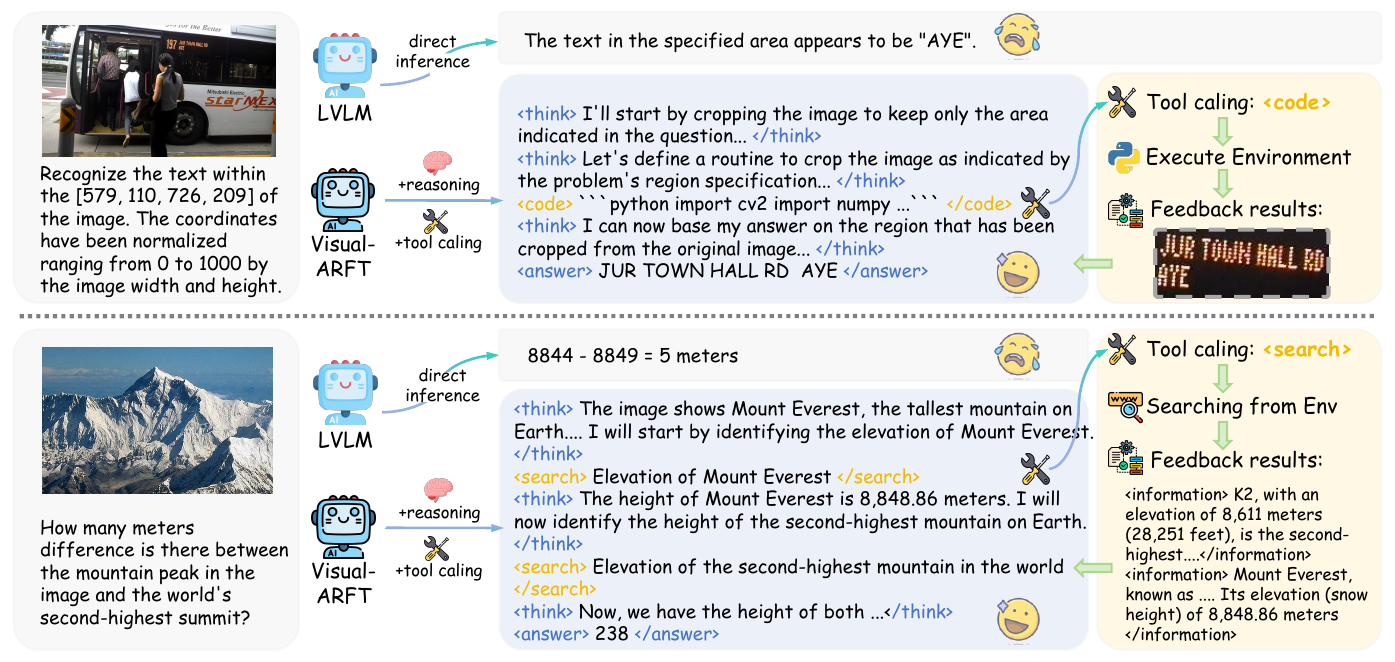}
   \vspace{-12pt}
    \caption{\small The benefits of our \textbf{Visual} \textbf{A}gentic \textbf{R}einforcement \textbf{F}ine-\textbf{T}uning (\methodname) to perform complex multi-modal reasoning tasks, such as (\textbf{top}) write and execute python code to accurately read text within a specified image region and (\textbf{bottom}) use internet search to answer a multi-hop question.}
    \vspace{-12pt}
    \label{fig:tesear}
\end{figure}

\section{Introduction}
Recent advances in Large Reasoning Models (LRMs) have given rise to a new generation of agentic systems—models that can reason, plan, and interact with external tools to solve complex tasks. Among them, OpenAI’s o3~\cite{OpenAI-o3} exemplifies a major leap forward by demonstrating native support for tool-augmented reasoning across both textual and visual modalities. These capabilities mark a shift from static, single-turn inference to dynamic, multi-step decision-making, enabling models to browse the web~\cite{li2025webthinker}, execute code~\cite{li2025torl}, and manipulate images \cite{OpenAI-o3} to complete real-world tasks.

While supervised fine-tuning has been the mainstream approach for enabling tool use in LLMs, it typically relies on curated demonstrations or handcrafted trajectories, which are costly to scale and difficult to generalize. A growing trend, however, is the use of Reinforcement Fine-Tuning (RFT) to train an agentic system. This shift is exemplified by OpenAI-o1~\cite{jaech2024openaio1}, which highlights RFT as a key technique for rapidly adapting reasoning models to new domains using only a small number of examples. Although the exact details of o1 remain proprietary, DeepSeek-R1~\cite{guo2025deepseek-r1} has shown that verifiable reward signals~\cite{lambert2024ttulu, team2025kimi}, derived from rule-based correctness checks rather than learned reward models~\cite{ouyang2022training, liu2024skywork, zang2025internlm}, can serve as an effective supervisory signal during RFT. 

Recent works such as Search-R1~\cite{jin2025search-r1} and ReTool~\cite{feng2025retool} have successfully applied Reinforcement Fine-Tuning (RFT) to agentic learning, enabling models to acquire tool-use abilities through verifiable reward. These approaches demonstrate promising results in tasks such as QA and mathematical reasoning, highlighting the effectiveness of RFT in teaching models to act and reason within tool-augmented environments. However, current efforts remain largely confined to language-only settings, and the potential of RFT in multimodal agentic reasoning, especially involving visual inputs and tool execution, remains underexplored.

To address this gap, we propose \textbf{V}isual \textbf{A}gentic \textbf{R}einforcement \textbf{F}ine-\textbf{T}uning (\methodname), an effective framework that equips Large Vision-Language Models (LVLMs) with agentic reasoning and tool-use abilities.
Unlike previous approaches that rely on manually designed prompts or massive supervised datasets, \methodname applies a reward-driven training strategy to teach LVLMs to reason, decompose tasks, and invoke tools when needed. We design a set of verifiable rewards specifically tailored for agentic reasoning and tool-use behaviors. These rewards provide explicit supervision signals that guide the direction of policy optimization~\cite{grpo,dpo,ppo}, enabling the model to learn effective strategies for tool invocation and reasoning. In detail, we adopt the Group Relative Policy Optimization (GRPO)~\cite{grpo} algorithm to update the model’s policy based on the reward feedback.

We focus on two challenging multimodal settings: agentic search and agentic coding, where the model must either (a) plan, decompose the original task, and retrieve information from external sources to answer complex multimodal multi-hop VQA questions, or (b) reason about the task, write and execute code to process the image, and solve challenging visual question answering problems.
As illustrated by the examples in Fig.~\ref{fig:tesear}, \methodname successfully completes complex multimodal tasks by leveraging its strong capabilities in task analysis, step-by-step reasoning, and tool invocation.

To facilitate training and evaluation, we introduce the \textbf{M}ultimodal \textbf{A}gentic \textbf{T}ool Bench (\benchname), which includes \searchname and \codename. \benchname is designed to measure agentic reasoning accuracy, tool execution correctness, and multimodal generalization. In constructing the dataset, we adopt different strategies: \searchname is built through manual annotation, while \codename is generated using an automated construction pipeline. To ensure the overall quality of the \benchname, all examples are manually reviewed and filtered after construction, resulting in a total of 350 high-quality examples.

Our experiments demonstrate that \methodname enables LVLMs like Qwen2.5-VL-3B/7B~\cite{bai2025qwen25vl} to achieve state-of-the-art results on both our proposed multimodal agentic benchmark (\codename, \searchname) and existing multi-hop QA benchmarks, while requiring only minimal annotated training data. Notably, \methodname outperforms GPT-4o on the \codename task using only a 3B base model. On the same benchmark, the 7B model achieves a significant improvement over the baseline, with a gain of +18.56\% F1 and +13.00\% EM. On the \searchname task, the 7B model also demonstrates strong performance, outperforming the baseline by +10.28\% F1 and +8.66\% EM. Furthermore, when evaluated on out-of-domain multi-hop QA benchmarks, \methodname continues to deliver substantial performance gains, highlighting its strong generalization capability across reasoning tasks.

In summary, our key contributions are as follows:

\noindent \textbf{(1)} We propose Visual Agentic Reinforcement Fine-Tuning (\methodname), an effective framework that equips LVLMs with agentic capabilities, including task planning, reasoning, and tool use (write and execute Python code, search the internet).

\noindent \textbf{(2)} We design modular verifiable rewards tailored to agentic reasoning and tool-invocation behaviors in both searching and coding tasks. These rewards allow the model to learn structured, interpretable behaviors without requiring preference data or learned reward models.

\noindent \textbf{(3)} We introduce the Multimodal Agentic Tool Bench (\benchname), consisting of \searchname and \codename, specifically constructed to evaluate multimodal agentic reasoning and tool-use capacity. \benchname includes high-quality, carefully annotated examples and diverse task settings.

\noindent \textbf{(4)} We conduct extensive experiments on both in-domain and out-of-domain benchmarks, where \methodname significantly outperforms baseline models. \methodname provides a promising technical path toward building powerful multimodal agent systems.


%% file: sec/2_related_work.tex
\section{Related Work} \label{sec_2_related_work}

\noindent \textbf{Agentic Reasoning and Tool Use.} Recent advances in large language models (LLMs) \cite{OpenAI-o3,hurst2024gpt4o, team2023gemini, wang2024qwen2vl,li2024llavaov,xcomposer2.5,liu2024deepseekv3} have driven the development of agentic systems—models capable of planning, reasoning, and interacting with external tools to solve complex tasks. While much prior work has focused on language-only agentic behaviors, such as function calling and API execution\cite{qian2025toolrl, zhang2025nemotron}, relatively little attention has been paid to multimodal agentic capabilities, where models must interpret visual inputs and invoke tools accordingly. Proprietary systems like OpenAI-o3~\citep{OpenAI-o3} demonstrate native support for web search and image manipulation, but open-source research still lacks standardized benchmarks and training methods for tool-augmented multimodal reasoning. Moreover, most existing datasets assume static reasoning, without multi-step decision-making or dynamic tool use, limiting their relevance for agentic training. Our work addresses these gaps by proposing \methodname, a unified framework for multimodal agentic training, and introducing the Multimodal Agentic Tool Bench (\benchname) to evaluate tool use and visual reasoning in both searching and coding tasks.

\noindent \textbf{Reinforcement Learning.} To move beyond the limitations of supervised fine-tuning, recent research has applied reinforcement learning (RL) with verifiable rewards to strengthen the reasoning ability~\cite{liu2025visualrft}. Methods like GRPO~\cite{grpo} simplify training with rule-based verifiable reward have shown success in math and coding tasks. In parallel, tool-augmented reasoning has gained attention, with models leveraging search~\cite{goldie2025synthetic, sun2025zerosearch, jin2025search-r1, song2025r1-searcher, researcher}, code execution~\cite{feng2025retool, li2025torl, liao2024mario}, or APIs to complete complex tasks. However, many tool-use methods rely on handcrafted prompts or large amount of data, and often decouple tool execution from the reasoning loop. Reinforcement learning has shown promise in improving tool-based reasoning in language models, particularly for tasks like search~\cite{jin2025search-r1, sun2025zerosearch, li2025search-o1,researcher} and math problem solving~\cite{singh2025agentic-RL-math}. However, its application to agentic reasoning in multimodal models has received far less attention. Our work addresses this gap by applying verifiable reward–based RL to teach LVLMs to plan, invoke tools, and solve complex multimodal tasks.

%% file: sec/3_methodology.tex
\section{Visual Agentic Reinforcement Fine-Tuning (Visual ARFT)} \label{sec_3_methodology}
\subsection{Preliminary}

\noindent \textbf{Reinforcement Learning with Verifiable Rewards.} Reinforcement Learning with Verifiable Rewards (RLVR)~\cite{guo2025deepseek-r1, lambert2024ttulu, team2025kimi} is a training paradigm aimed at improving language models on tasks where correctness can be directly and objectively verified, such as code generation or mathematical reasoning. Unlike traditional RLHF methods~\cite{ouyang2022training,dpo, ppo, sun2023aligning, zhou2024aligning, llavarlhf, yu2024rlhfv, yu2024rlaif, liu2024mia}, which rely on human-annotated preferences and learned reward models, RLVR bypasses subjective feedback by employing a deterministic, programmatic reward signal.

The central idea behind RLVR is to define a reward function that determines whether a model’s output $o$ matches the ground-truth answer for a given input $q$. This allows the model to receive precise supervision without introducing uncertainty from noisy or inconsistent reward models.

Formally, the training objective in RLVR can be expressed as maximizing the expected reward, while simultaneously constraining the learned policy $\pi_\theta$ to remain close to a reference policy $\pi_{\text{ref}}$:
\begin{equation}
\max_{\theta}\mathbb{E}_{o \sim \pi_\theta(o|q)}[R_{RLVR(q, o)}]  = \left[ R(q, o) \right] - \beta \cdot \text{KL} \left( \pi_\theta(o|q) || \pi_{\text{ref}}(o|q) \right),
\end{equation}
where $\beta$ is a regularization coefficient controlling the trade-off between reward maximization and policy stability. The verifiable reward function $R(q, o)$ is defined as:
\begin{equation}
R(q, o) =
\mathbb{I} \left[ o = \text{ground-truth}(q) \right],
\end{equation}
where $\mathbb{I}[\cdot]$ is the indicator function returning 1 if the prediction is exactly correct and 0 otherwise.
This reward structure ensures that the learning signal aligns strictly with task-defined correctness, and enables efficient training in domains where outputs can be unambiguously validated.

\begin{figure}[t]
    \centering
   \includegraphics[width=.98\linewidth]{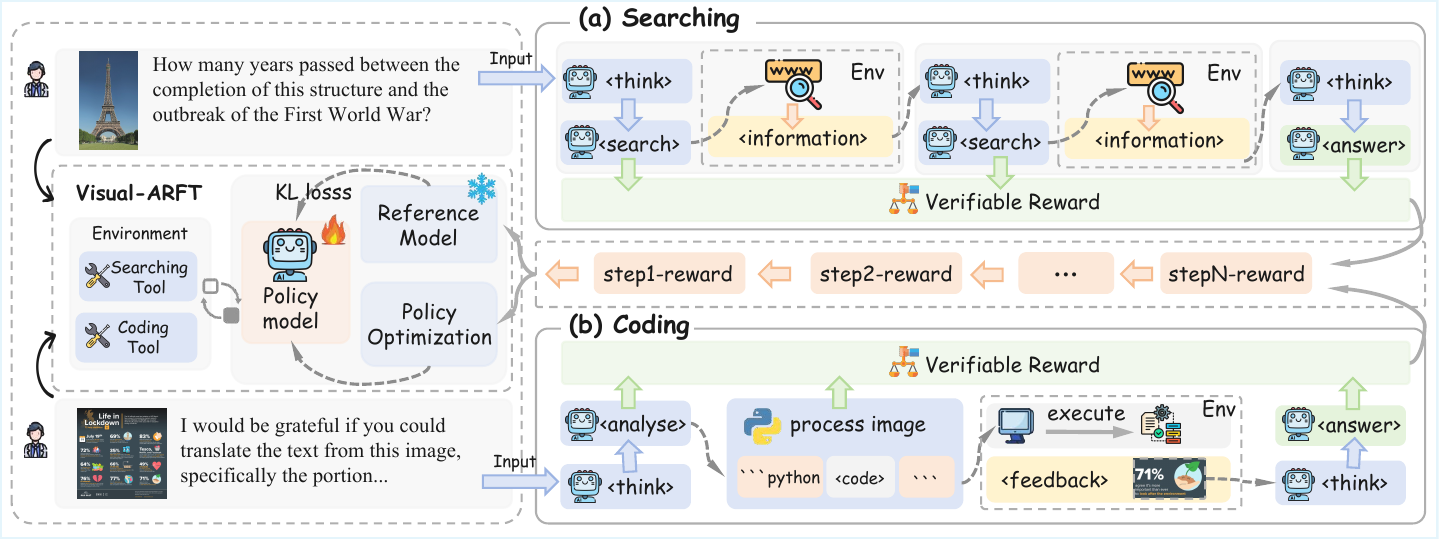}
   \vspace{-12pt}
    \caption{\textbf{\small Overview of \methodname.} We successfully empower LVLMs with multimodal agentic capabilities, including (a) agentic search and (b) agentic coding, enabling them to solve complex multimodal tasks through reasoning, decomposition, and tool interaction.}
    \vspace{-12pt}
    \label{fig:framework}
\end{figure}

\subsection{Framework of \methodname}

Fig.~\ref{fig:framework} illustrates the overall framework of \methodname, which is applied to two multimodal agentic scenarios: search and coding. Given multimodal inputs, the policy model $\pi_\theta$ generates a sequence of responses that include both intermediate reasoning steps (e.g., \texttt{<think>}) and action decisions, such as invoking a search tool or generating Python code to solve complex tasks.

In the search setting, the model is presented with complex multimodal multi-hop VQA queries that often involve multiple sub-questions and require external factual knowledge. Such queries are difficult to resolve using only in-context knowledge or Chain-of-Thought (CoT) prompting. Therefore, the model must be capable of reasoning, decomposing the original question into manageable sub-tasks, and invoking tools to retrieve relevant information from external sources. To address these challenges, \methodname trains the model to autonomously analyze and decompose multi-hop queries and solve them through iterative interactions with a web search engine.

In the agentic coding setting, the input image may suffer from visual degradation or contain redundant content, where only specific regions are relevant to the question. These conditions impose a significant burden on the model’s visual understanding capabilities. \methodname is designed to handle this by training the model to actively generate executable code conditioned on the visual question. The model submits the code to an external environment to preprocess the image (e.g., rotation, cropping), thereby extracting useful information and enabling accurate question answering.

During training, the model typically requires multiple reasoning steps or repeated tool invocations to complete a single task. To guide this process, we design a rule-based verifiable reward that evaluates both the model’s tool usage and final answers. Rewards are assigned at each step and provide learning signals for policy optimization. We adopt GRPO~\cite{grpo} for training, incorporating a KL divergence term to prevent the updated policy from drifting too far from the reference model. This regularization mitigates reward hacking and promotes both training stability and generalization.

\subsubsection{Reward Design}\label{sec:methoda_reward}
The design of reward functions plays a critical role in determining the effectiveness of reinforcement learning. Verifiable rewards use simple correctness checks to determine whether an answer is right or wrong, without relying on learned reward models or human feedback. In particular, the minimalistic design of exact match–based verifiable rewards has been shown by DeepSeek-R1 to significantly enhance model reasoning performance, demonstrating strong scalability and general applicability.

\paragraph{Format Reward.} In \methodname, we require the model to generate outputs that strictly follow a predefined format. This form of structured reward encourages the model to engage in step-by-step reasoning and to invoke tools as instructed, thereby improving the reliability of tool execution. In both the Coding and Searching tasks, the model’s reasoning process is enclosed within `<think></think>' tags. When the model needs to invoke a search or code execution tool, it outputs a `<search></search>' or  `<code></code>' tag, respectively. The content within each tool tag, either the search query or the generated code, is then extracted and used to interact with the external environment:
\begin{equation}
R_{\text{format}}(o) = \mathbb{I}\left[o \text{ contains valid tags} \right].
\end{equation}
\noindent \textbf{Accuracy Rewards.} The accuracy reward in Visual-RFT is composed in a modular fashion. For the final answers in both the searching and coding tasks, we use the F1 score as the reward to evaluate answer quality ($R_{F1}$). F1 rewards offer greater tolerance than exact match, providing smoother and more informative learning signals. This design better captures the fluency and variability of natural language responses, and contributes to more stable reinforcement learning.

For evaluating the effectiveness of search tool usage, we compute the semantic similarity between the model-generated search query and the ground-truth query using a Sentence Transformer. Compared to F1-based rewards, semantic similarity rewards ($R_{sem}$) are more robust to surface-level variations and better reflect whether the model captures the true intent of the retrieval objective. This is especially important for open-ended or paraphrased queries, where lexical overlap may be low but semantic equivalence still holds.

For the coding component, we assign a reward of 1 to all outputs that fall within executable code blocks, without directly supervising the content of the generated code. Instead of evaluating the correctness of the code itself, we allow the model to freely synthesize code based on its own reasoning. This design encourages flexibility and open-ended problem-solving.
In preliminary experiments, we observed that applying strict correctness-based rewards to code generation led the model to converge on a small set of deterministic solutions, which in turn reduced its ability to generalize across diverse visual reasoning scenarios. By removing content-level supervision on code, we promote broader exploration and maintain the agentic nature of decision making.
\begin{equation}
R_{\text{acc}}(q, o) = 
\begin{cases}
R_{\text{F1}}(o_{\text{ans}}, a), & \text{if } o \text{ is the final answer (}<\texttt{answer}>\text{)}, \\
R_{\text{sem}}(o_{\text{search}}, s), & \text{if } o \text{ is a search query (}<\texttt{search}>\text{)}, \\
1, & \text{if } o \text{ is a code block (}<\texttt{code}>\text{).}
\end{cases}
\end{equation}
\noindent \textbf{Total Rewards.} The total reward used in Visual-ARFT combines two components: a format reward that encourages the model to follow the required output structure, and an accuracy reward that evaluates the quality of answers and tool usage. The format reward ensures proper tagging for reasoning and tool invocation, while the accuracy reward provides feedback on the final answer and search content. The overall reward function is defined as:
\begin{equation}
R_{\text{total}}(q, o) = R_{\text{format}}(o) + R_{\text{acc}}(q, o).
\end{equation}

%% file: sec/4_benchmark.tex
\section{Multimodal Agentic Tool (\benchname) Benchmark} \label{sec:mat}
Given that existing benchmarks for agentic search are predominantly language-only~\cite{bfcl3,ho2020constructing2wiki,yang2018hotpotqa,trivedi2022musique,press2022measuringbamboodle}, and that there is a lack of standardized evaluation protocols for multimodal agentic coding tasks, we construct and annotate a new benchmark to fill this gap.
We refer to the full dataset as the Multimodal Agentic Tool Bench (\benchname), which comprises two sub-settings: \searchname and \codename.
Specifically, we create a set of multimodal multi-hop VQA examples for evaluating agentic searching capabilities, and another set of VQA examples involving image distortions that require tool-based image processing for agentic coding.

\begin{figure}[t]
    \centering
   \includegraphics[width=.98\linewidth]{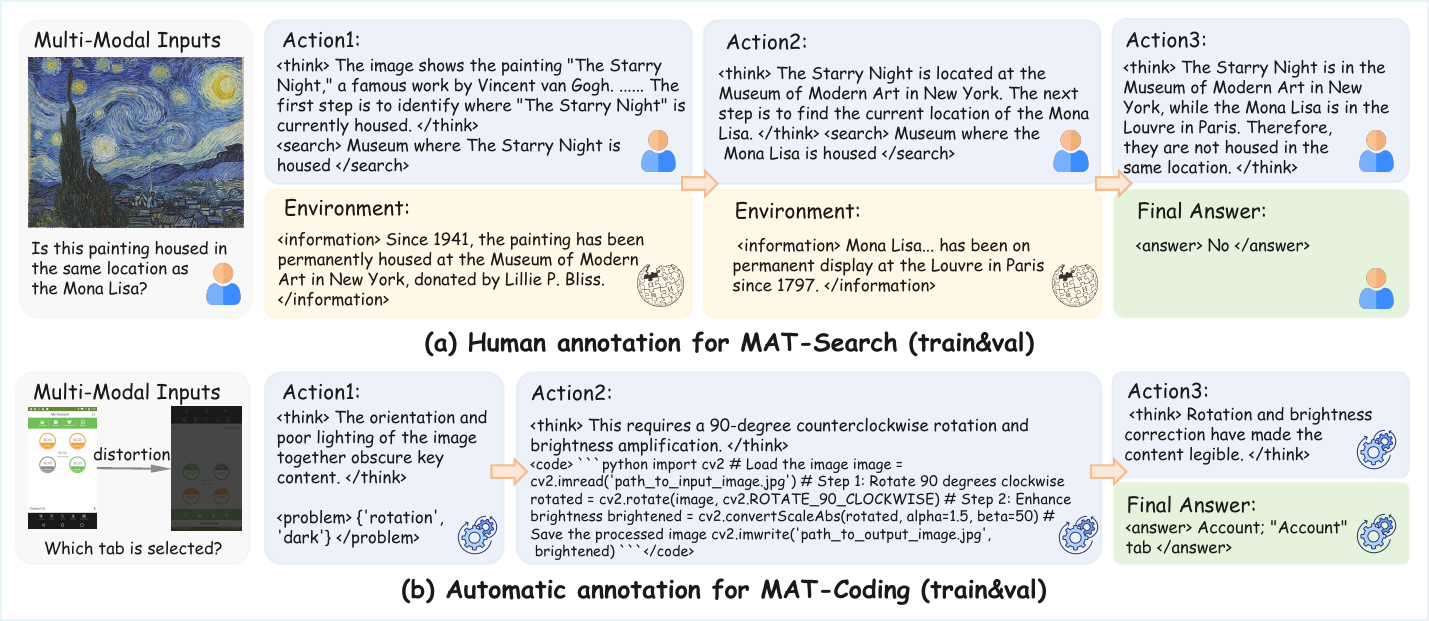}
   \vspace{-12pt}
    \caption{\small \textbf{Data Annotation Pipeline} of our proposed Multimodal Agentic Tool Bench (\benchname): (a) \searchname, a manually annotated and verified dataset for agentic search, and (b) \codename, an automatically generated dataset for agentic coding with a structured pipeline.}
    \vspace{-12pt}
    \label{fig:data_annotation}
\end{figure}

\noindent \textbf{\searchname.}
For the agentic search task, as illustrated in Fig.~\ref{fig:data_annotation} (a), we manually construct a multimodal multi-hop VQA dataset. Specifically, we build the \searchname benchmark by annotating 150 high-quality multimodal multi-hop VQA examples. The questions in \searchname vary in difficulty and require different levels of reasoning depth—more complex queries involve more inference steps and factual knowledge, challenging the model's ability to handle composite problems and retrieve relevant external information. Each example is carefully crafted and verified by human annotators to ensure clarity, consistency, and suitability for evaluating agentic reasoning and tool use.

\noindent \textbf{\codename.} For the agentic coding task, as shown in Fig.~\ref{fig:data_annotation} (b), we design an automated data annotation pipeline. Starting from existing VQA examples, we apply various types of image distortions—including rotation, brightness adjustment, blurring, noise injection, and hybrid distortions—to increase the difficulty of visual understanding. Under this setup, directly reasoning over the processed image poses substantial challenges for the model. We annotate 200 test samples to evaluate the model’s ability to complete the image distortion tasks. While the model may attempt to answer questions directly from the processed image, it must overcome various forms of visual noise and irrelevant content. In contrast, advanced agentic systems, such as OpenAI-o3 and our \methodname, can invoke code, manipulate the image, and then utilize the processed result to more effectively complete the VQA task. Further details on data types and sources are provided in the \cref{sec:appendix_datas}.

\noindent \textbf{\benchname Training Data.}
While OpenAI-o3 demonstrates strong capabilities in agentic search and coding, existing open-source multimodal models have not been specifically trained on such tasks. Therefore, we construct a training dataset for \methodname. Given the simplicity and regularity of the search task format, we carefully annotate \textbf{20} multimodal QA examples requiring multi-step reasoning, each paired with a well-defined chain of thought and final answer, which are used to train \methodname in the agentic search setting.
In contrast, the agentic coding task involves more complex workflows, including the model's autonomous generation of code.
To support this, we construct a training set of \textbf{1,200} examples to enable effective learning under this setting.

%% file: sec/4_experiments.tex
\vspace{-12pt}
\section{Experiments}
\vspace{-4pt}

\noindent \textbf{Datasets and Evaluation Metrics.} \label{sec: exp_data_detail}
We evaluate our method on three settings: \textbf{(1)} agentic coding on our proposed \codename, \textbf{(2)} agentic search on our proposed \searchname.;
\textbf{(3)} generalize on existing multihop QA benchmarks, including 2WikiMultihopQA \cite{ho2020constructing2wiki}, HotpotQA \cite{yang2018hotpotqa}, MuSiQue \cite{trivedi2022musique}, and Bamboogle \cite{press2022measuringbamboodle}.
For all settings, we use F1 and Exact Match (EM) metrics, where F1 captures token-level overlap and EM requires an exact string match with the ground-truth answer.

\noindent \textbf{Implementation Details.}  \label{sec: exp_imple_detail}
We apply \methodname on Qwen2.5-VL-3B and 7B.
Training is performed on 8 GPUs using the GRPO algorithm, with 8 sampled generations used for policy gradient estimation per update. 
For agentic search, we use the Serper API as the retrieval tool, providing access to Google's search engine.
This API-based approach avoids the overhead of building a full RAG system and better reflects how agents interact with external environments in real-world settings.

\subsection{Main Results}

\noindent \textbf{Results on \codename.} We evaluate our method on the \codename benchmark. Compared to the base Qwen2.5-VL-3B/7B models, \methodname enables models to autonomously analyze distorted or misaligned images—including both degraded and crop-requiring inputs—generate preprocessing code, and complete the QA pipeline, showcasing their agentic capabilities in reasoning, task decomposition, and tool use.

As shown in Tab.\ref{tab:coding_bench}, the Qwen2.5-VL-7B model achieves substantial improvements with \methodname, gaining +18.56 F1 and +13.00 EM on average. The gains are especially notable on the Hard subset, where base models struggle due to limited reasoning and no tool-use ability. With \methodname, the model invokes code-based tools (e.g., repair, crop), resulting in strong performance gains. Example cases are shown in Fig.\ref{fig:case_mat_infere}. The 3B model also benefits from \methodname, with consistent F1 improvements. EM remains stable due to strict formatting requirements, but the F1 gains reflect improved task completion ability enabled by agentic reasoning with tool.

We also observe that OpenAI-o3, with native tool-use ability, achieves the highest performance, surpassing GPT-4o. Its ability to reason and manipulate images via code highlights the strength of agentic multimodal models. In summary, \methodname significantly improves model performance on \codename, especially for harder tasks. By introducing agentic capabilities, it enables LVLMs to reason, invoke tools, and solve complex visual problems effectively.

\begin{table}[t]
\caption{
\small
\textbf{Results of \benchname.} We conducted experiments on \benchname, including \codename and \searchname, and the table presents the evaluation results of several open-source and proprietary models on our benchmark. We use F1 score and Exact Match (EM) to evaluate model performance.}
\vspace{-12pt}
\label{tab:coding_bench}
\begin{center}
\setlength{\tabcolsep}{2pt}
\renewcommand{\arraystretch}{1.1}
\scalebox{0.75}{
\begin{tabular}{l|c|
cc|cc|
cc|cc|cc|cc}
\toprule
\multirow{3}{*}{\textbf{Models}} & 
\multirow{3}{*}{\shortstack{\textbf{Reasoning} \\ \textbf{with} \\ \textbf{Tools} }} &
\multicolumn{6}{c|}{\textbf{\codename}} &
\multicolumn{6}{c}{\textbf{\searchname}} \\
\cmidrule(lr){3-8} \cmidrule(lr){9-14}
& &
\multicolumn{2}{c|}{\textbf{Simple}} &
\multicolumn{2}{c|}{\textbf{Hard}} &
\multicolumn{2}{c|}{\textbf{Avg}} &
\multicolumn{2}{c|}{\textbf{Simple}} &
\multicolumn{2}{c|}{\textbf{Hard}} &
\multicolumn{2}{c}{\textbf{Avg}} \\
\cmidrule(lr){3-4} \cmidrule(lr){5-6} \cmidrule(lr){7-8} \cmidrule(lr){9-10} \cmidrule(lr){11-12} \cmidrule(lr){13-14}
& &
\textit{F1} & \textit{EM} 
& \textit{F1} & \textit{EM}
& \textit{F1} & \textit{EM}
& \textit{F1} & \textit{EM}
& \textit{F1} & \textit{EM}
& \textit{F1} & \textit{EM}\\
\midrule
GPT-4o~\cite{hurst2024gpt4o}  & \xmark  & 47.12 & 38.57 & 27.57 & 15.38 & 34.41 & 23.5 & 68.55 & 61.33 & 53.61 & 42.67 & 61.08 & 52.00 \\
OpenAI-o3~\cite{OpenAI-o3}  & \checkmark  & 70.38 & 65.38 & 75.00 & 70.59 & 72.99 & 68.33 & 79.72 & 70.67 & 63.74 & 52.00 & 71.73 & 61.33\\
\midrule
LLaVa-v1.5-7B~\cite{llava1.5}  & \xmark & 19.50 & 12.86 & 9.30 & 5.38 & 12.87 & 8.00 & 56.55 & 52.00 & 30.32 & 25.33 & 43.44 & 38.67 \\
LLaVa-Next-7B~\cite{li2024llavanext}  & \xmark & 30.78 & 17.14 & 17.11 & 10.00 & 21.89 & 12.5 & 63.27 & 56.00 & 38.75 & 29.33 & 51.01 & 42.67 \\
LLaVa-OneVision-7B~\cite{li2024llavaov}  & \xmark & 39.86 & 28.57 & 16.05 & 11.54 & 24.38 & 17.5 &61.78 & 54.67 & 31.66 & 26.67 & 46.72 & 40.67 \\
Xcomposer2.5~\cite{xcomposer2.5}  & \xmark & 36.06 & 22.86 & 19.90 & 10.77 & 25.56 & 15.0 & 60.16 & 54.67 & 31.93 & 28.00 & 46.04 & 41.33 \\
InternVL2.5-8B~\cite{internvl2.5}  & \xmark & 39.48 & 28.57 & 26.62 & 13.85 & 31.12 & 19.00 & 61.72 & 53.33 & 41.69 & 33.33 &  51.70 & 43.33 \\
\midrule
Qwen2.5-VL-3B~\cite{bai2025qwen25vl}  & \xmark & 46.29 & 35.71 & 17.98 & 13.85 & 27.89 & 21.50 & 57.54 & 50.67 & 33.11 & 26.67 & 45.32 & 38.67 \\
\rowcolor[HTML]{DAEFF9} \quad + \methodname & \checkmark & 49.78 & 40.00 & 28.42 & 13.08 & 35.90 & 22.50 & 56.41 & 50.67 & 45.55 & 36.00 & 50.98 & 43.33 \\
\quad $\Delta$ & -- & \hgreen{+3.49} & \hgreen{+4.29} & \hgreen{+10.44} & \hblue{-0.78} & \hgreen{+8.01} & \hgreen{+1.0} & \hblue{-1.13} & \hgreen{+0.0} & \hgreen{+12.44} & \hgreen{+9.33} & \hgreen{+5.66} & \hgreen{+4.66} \\
\midrule
Qwen2.5-VL-7B~\cite{bai2025qwen25vl}  & \xmark & 55.23 & 40.00 & 19.67 & 11.54 & 32.12 & 21.50 & 67.40 & 61.33 & 39.59 & 32.00 & 53.49 & 46.67 \\
\rowcolor[HTML]{DAEFF9} \quad + \methodname & \checkmark & 60.10 & 51.43 & 45.60 & 25.38 & 50.68 & 34.50 & 71.78 & 66.67 & 55.77 & 44.00 & 63.77 & 55.33 \\
\quad $\Delta$ & -- & \hgreen{+4.87} & \hgreen{+11.43} & \hgreen{+25.93} & \hgreen{+13.84} & \hgreen{+18.56} & \hgreen{+13.00} & \hgreen{+4.38} & \hgreen{+5.37} & \hgreen{+16.18} & \hgreen{+12.00} & \hgreen{+10.28} & \hgreen{+8.66} \\
\bottomrule
\end{tabular}
}
\end{center}
\vspace{-20pt}
\end{table}


\noindent \textbf{Results on \searchname.} We compare our proposed method against several open-source and proprietary models on \searchname. As shown in Tab.~\ref{tab:coding_bench}, \methodname improves the average F1 score by 5.66\% and the EM score by 4.66\% on Qwen2.5-VL-3B. On Qwen2.5-VL-7B, the improvements are even more significant, with F1 and EM gains of 10.28\% and 8.66\%, respectively.

These gains highlight the effectiveness of reinforcement fine-tuning with verifiable rewards in equipping the model with structured multimodal reasoning and tool-use capabilities. \methodname enables step-by-step planning and dynamic evidence acquisition, which are crucial for solving complex multi-hop queries. Additionally, we include comparisons with traditional approaches such as RAG and CoT in our ablation study (\cref{sec:ablation}) to further validate the benefits of our method.

Powered by efficient tool usage and task decomposition capabilities, \methodname outperforms GPT-4o on the \searchname benchmark. Similar to \codename, we also note that OpenAI's o3, with its inherently strong reasoning ability, still demonstrates superior performance on \searchname and outperforms all open-source models. This underscores the need for further research into the agentic capabilities of open-source multi-modal models.

\begin{table}[t]
\caption{
\small \textbf{Results on Existing Text-Only Multihop QA Benchmarks.} Our \methodname outperforms strong baselines such as Search-o1 \cite{li2025search-o1}, Search-R1 \cite{jin2025search-r1}, and ZeroSearch \cite{sun2025zerosearch}.}
\label{tab:multihopqa}
\vspace{-8pt}
\begin{center}
\setlength{\tabcolsep}{2pt}
\renewcommand{\arraystretch}{1.1}
\scalebox{0.80}{
\begin{tabular}{l|c|
cc|cc|cc|cc|cc}
\toprule
\multirow{2}{*}{\textbf{Models}} & 
\multirow{2}{*}{\shortstack{\textbf{Reasoning} \\ \textbf{with Tools}}} &
\multicolumn{2}{c|}{\textbf{2Wiki} \cite{ho2020constructing2wiki}} &
\multicolumn{2}{c|}{\textbf{HotpotQA} \cite{yang2018hotpotqa}} &
\multicolumn{2}{c|}{\textbf{MuSiQue} \cite{trivedi2022musique}} &
\multicolumn{2}{c|}{\textbf{Bamboogle} \cite{press2022measuringbamboodle}} &
\multicolumn{2}{c}{\textbf{Avg}} \\
\cmidrule(lr){3-4} \cmidrule(lr){5-6} \cmidrule(lr){7-8} \cmidrule(lr){9-10} \cmidrule(lr){11-12}
& &
\textit{F1} & \textit{EM} 
& \textit{F1} & \textit{EM}
& \textit{F1} & \textit{EM}
& \textit{F1} & \textit{EM}
& \textit{F1} & \textit{EM}\\
\midrule
\multicolumn{10}{l}{\textbf{\textit{Qwen-2.5-7B-Instruct}}~\cite{bai2025qwen25vl}} \\
\quad + Direct Inference & \xmark  & -- & 25.00 & -- & 18.30 & -- & 3.10 & -- & 12.00 & -- & 14.60\\
\quad + CoT & \xmark  & -- & 11.00 & -- & 9.20 & -- & 2.20 & -- & 23.20 & -- & 11.40\\
\quad + IRCoT & \xmark  & -- & 14.90 & -- & 13.30 & -- & 7.20 & -- & 22.40 & -- & 14.45\\
\quad + RAG & \xmark  & -- & 23.50 & -- & 29.90 & -- & 5.80 & -- & 20.80 & -- & 20.00\\
\midrule
\quad + Search-o1~\cite{li2025search-o1} & \checkmark  & -- & 17.60 & -- & 18.70 & -- & 5.80 & -- & 29.60 & -- & 17.93\\
\quad + Search-R1~\cite{jin2025search-r1}  & \checkmark  & -- & 41.40 & -- & \textbf{37.00} & -- & 14.60 & -- & 36.80 & -- & 32.45\\
\quad + ZeroSearch~\cite{sun2025zerosearch}  & \checkmark  & -- & 43.12 & -- & 29.21 & -- & \textbf{19.72} & -- & 35.20 & -- & 31.81\\
\midrule
\multicolumn{10}{l}{\textbf{\textit{Qwen2.5-VL-3B-Instruct}}~\cite{bai2025qwen25vl}} \\
\quad + Direct Inference  & \xmark & 31.60 & 26.40 & 22.90 & 15.84  & 9.64 & 2.36 & 10.82 & 5.60 & 18.74 & 12.55 \\
\quad + RAG  & \xmark & 35.14 & 26.92 & 35.08 & 23.67 & 15.08 & 7.61 & 28.37 & 19.20 & 28.42 & 19.35 \\
									
\rowcolor[HTML]{DAEFF9} \quad + \methodname & \checkmark & 49.77 & 39.13 & 41.33 & 30.10 & 19.10 & 11.58 & 49.01 & 38.40 & 39.80 & 29.80 \\
\quad $\Delta$ & -- & \hgreen{+18.17} & \hgreen{+12.73} & \hgreen{+18.43} & \hgreen{+14.26} & \hgreen{+9.46} & \hgreen{+9.22} & \hgreen{+38.19} & \hgreen{+32.80} & \hgreen{+21.06} & \hgreen{+17.25} \\
\midrule
\multicolumn{10}{l}{\textbf{\textit{Qwen2.5-VL-7B-Instruct}}~\cite{bai2025qwen25vl}} \\
\quad + Direct Inference  & \xmark & 27.90 & 22.15 & 25.35 & 17.35 & 9.45 & 2.61 & 14.07 & 5.60 & 19.19 & 11.93 \\
\quad + RAG  & \xmark & 31.58 & 21.74 & 36.33 & 24.09 & 15.36 & 7.61 & 27.60 & 19.20 & 19.82 & 12.73\\
\rowcolor[HTML]{DAEFF9} \quad + \methodname & \checkmark & \textbf{63.09} & \textbf{51.99} & \textbf{48.00} & 36.48 & \textbf{22.71} & 14.07 & \textbf{60.15} & \textbf{48.80} & \textbf{48.49} & \textbf{37.84} \\
\quad $\Delta$ & -- & \hgreen{+35.19} & \hgreen{+29.84} & \hgreen{+22.65} & \hgreen{+19.13} & \hgreen{+13.26} & \hgreen{+11.46} & \hgreen{+46.08} & \hgreen{+43.20} & \hgreen{+29.30} & \hgreen{+25.91} \\
\bottomrule
\end{tabular}
}
\end{center}
\vspace{-24pt}
\end{table}

\noindent \textbf{Results on Existing Multi-Hop QA Benchmarks.} To further evaluate the generalization ability of \methodname, we conducted comprehensive experiments on several existing text-only multi-hop QA benchmarks, including 2WikiMultihopQA \cite{ho2020constructing2wiki}, HotpotQA \cite{yang2018hotpotqa}, MuSiQue \cite{trivedi2022musique}, and Bamboogle \cite{press2022measuringbamboodle}. 

Here we apply \methodname on Qwen2.5-VL-3B/-7B models using 20 manually annotated multimodal multi-hop VQA training examples (see \cref{sec:mat}).
Since our training is performed on VQA-style data involving both visual and textual modalities, there exists a clear modality and input-type gap between our training set and these language-only QA benchmarks \cite{ho2020constructing2wiki,yang2018hotpotqa,trivedi2022musique,press2022measuringbamboodle}. As such, the \textbf{out-of-domain} evaluations provide a strong testbed for validating the cross-modal generalization capability of \methodname.

Tab.~\ref{tab:multihopqa} presents a comparison with various baselines built on Qwen2.5-3B/7B, covering traditional approaches such as Chain-of-Thought (CoT) and Retrieval-Augmented Generation (RAG), as well as reinforcement learning-based methods like Search-R1 and ZeroSearch. As shown in Tab.~\ref{tab:multihopqa}, \methodname yields substantial performance gains on these out-of-domain multi-hop QA benchmarks. The Qwen2.5-VL-3B model improves by +21.06\% F1 and +17.25\% EM over the baseline. The 7B model further surpasses all baselines, with average EM gains of 25.91\% over direct inference. These improvements stem from \methodname’s ability to perform task decomposition and dynamic tool use, enabling step-by-step reasoning with flexible interaction with external information. Despite using only 20 training examples, \methodname achieves superior data efficiency and strong generalization across unseen domains. Moreover, \methodname is built on LVLM and outperforms several LMM-based methods, showcasing strong modality transfer and task generalization capabilities.

\subsection{Ablation Studies and Visualization Results} \label{sec:ablation}
\noindent \textbf{Reward Design.} To assess our reward design (\cref{sec:methoda_reward}), we replace the F1 score-based reward
\begin{wraptable}{r}{6.5cm}
\begin{minipage}{0.48\textwidth}
\caption{\small \textbf{Ablation Study} on the reward design.}
\label{tab:ablation_study}
\begin{center}
\adjustbox{width=.99\linewidth}{
\begin{tabular}{l|c|cc|cc|cc}
\toprule
\multirow{2}{*}{Task} &
\multirow{2}{*}{Reward} & 
\multicolumn{2}{c|}{\textbf{Simple}} &
\multicolumn{2}{c|}{\textbf{Hard}} &
\multicolumn{2}{c}{\textbf{Avg}} \\
\cmidrule(lr){3-4} \cmidrule(lr){5-6} \cmidrule(lr){7-8}
& & \textit{F1} & \textit{EM} & \textit{F1} & \textit{EM} & \textit{F1} & \textit{EM} \\
\midrule
MAT & EM & 53.50 & 45.71 & 43.00 & 23.85 & 46.68  & 31.50 \\
Coding & \cellcolor[HTML]{DAEFF9} F1 
  & \cellcolor[HTML]{DAEFF9} 60.10 & \cellcolor[HTML]{DAEFF9} 51.43 
  & \cellcolor[HTML]{DAEFF9} 45.60 & \cellcolor[HTML]{DAEFF9} 25.38 
  & \cellcolor[HTML]{DAEFF9} 50.68 & \cellcolor[HTML]{DAEFF9} 34.50 \\
\midrule

MAT & EM 
  & 59.23 & 53.33 
  & 53.37 & 41.33 
  & 56.30 & 47.33 \\
Search  & \cellcolor[HTML]{DAEFF9} F1 
  & \cellcolor[HTML]{DAEFF9} 71.78 & \cellcolor[HTML]{DAEFF9} 66.67 
  & \cellcolor[HTML]{DAEFF9} 55.77 & \cellcolor[HTML]{DAEFF9} 44.00 
  & \cellcolor[HTML]{DAEFF9} 63.77 & \cellcolor[HTML]{DAEFF9} 55.33 \\
\bottomrule
\end{tabular}
}
\end{center}
\end{minipage}
\hfill
\begin{minipage}{0.48\textwidth}
\caption{\small \textbf{Ablation Study} with Direct inference/CoT/RAG baselines on \searchname.}
\label{tab:ablation_study_cot_rag}
\begin{center}
\adjustbox{width=.99\linewidth}{
\begin{tabular}{l|c|cc|cc|cc}
\toprule
\multirow{2}{*}{Task} &
\multirow{2}{*}{Method} & 
\multicolumn{2}{c|}{\textbf{Simple}} &
\multicolumn{2}{c|}{\textbf{Hard}} &
\multicolumn{2}{c}{\textbf{Avg}} \\
\cmidrule(lr){3-4} \cmidrule(lr){5-6} \cmidrule(lr){7-8}
& & \textit{F1} & \textit{EM} & \textit{F1} & \textit{EM} & \textit{F1} & \textit{EM} \\
\midrule
Qwen2.5 & Direct & 57.54 & 50.67 & 33.11 & 26.67 & 45.32 & 38.67 \\
\quad -VL & CoT & 37.84 & 30.67 & 31.03 & 24.00 & 34.43 & 27.33 \\
\quad -3B & RAG & 49.42 & 45.33 & 39.07 & 32.00 & 44.25 & 38.67 \\
 & \methodname & \cellcolor[HTML]{DAEFF9} 56.41 & \cellcolor[HTML]{DAEFF9} 50.67 &
\cellcolor[HTML]{DAEFF9} 45.55 & \cellcolor[HTML]{DAEFF9} 36.00 & 
\cellcolor[HTML]{DAEFF9} 50.98 & \cellcolor[HTML]{DAEFF9} 43.33 \\
\midrule
Qwen2.5 & Direct & 67.40 & 61.33 & 39.59 & 32.00 & 53.49 & 46.67 \\
\quad -VL & CoT & 57.57 & 49.33 & 46.65 & 32.00 & 52.11 & 40.67 \\
\quad -7B & RAG & 59.14 & 56.00 & 42.44 & 36.00 & 50.79 & 46.00 \\
 & \methodname & \cellcolor[HTML]{DAEFF9} 71.78 & \cellcolor[HTML]{DAEFF9} 66.67 & 
\cellcolor[HTML]{DAEFF9} 55.77 & \cellcolor[HTML]{DAEFF9} 44.00 &
\cellcolor[HTML]{DAEFF9} 63.77 & \cellcolor[HTML]{DAEFF9} 55.33 \\
\bottomrule
\end{tabular}
}
\end{center}
\end{minipage}
\end{wraptable}
with an EM-based alternative for both agentic search and coding.
As shown in Tab.~\ref{tab:ablation_study}, EM-based training offers a slight improvement over the baseline, it consistently underperforms F1-based training.
This underscores the benefits of using the F1 score as a training signal: it provides smoother, more informative gradients by accounting for partial correctness, is more tolerant of linguistic variation, and facilitates stable training. Consequently, the F1 score-based reward is our default reward design.

\noindent \textbf{Comparison with RAG and CoT baselines.}
\methodname achieves a substantial performance improvement over the baseline on \searchname. To further evaluate model capabilities, we compare it with several traditional approaches, including Retrieval-Augmented Generation (RAG) and Chain-of-Thought (CoT). As shown in Tab.~\ref{tab:ablation_study_cot_rag}, CoT facilitates reasoning in multi-hop question answering by encouraging step-by-step inference.
However, it lacks access to external knowledge, limiting its ability to answer fact-based questions. In contrast, RAG can retrieve relevant external information but struggles with reasoning and problem decomposition, making it less effective on \searchname.
\methodname overcomes the limitations of both approaches by integrating strong reasoning, task decomposition, and tool-use abilities, significantly outperforming both RAG and CoT.

Additionally, we observe that methods based on CoT prompting or standard RAG not only fail to improve performance, but in some cases, lead to degradation. This suggests that for base models without task-specific adaptation, directly injecting external information—such as content retrieved via RAG—can be ineffective or even counterproductive due to information interference. In contrast, models trained with \methodname learn not only how to invoke tools, but also how to selectively utilize relevant external information to support complex reasoning.

\begin{figure}[t]
    \centering
   \includegraphics[width=.98\linewidth]{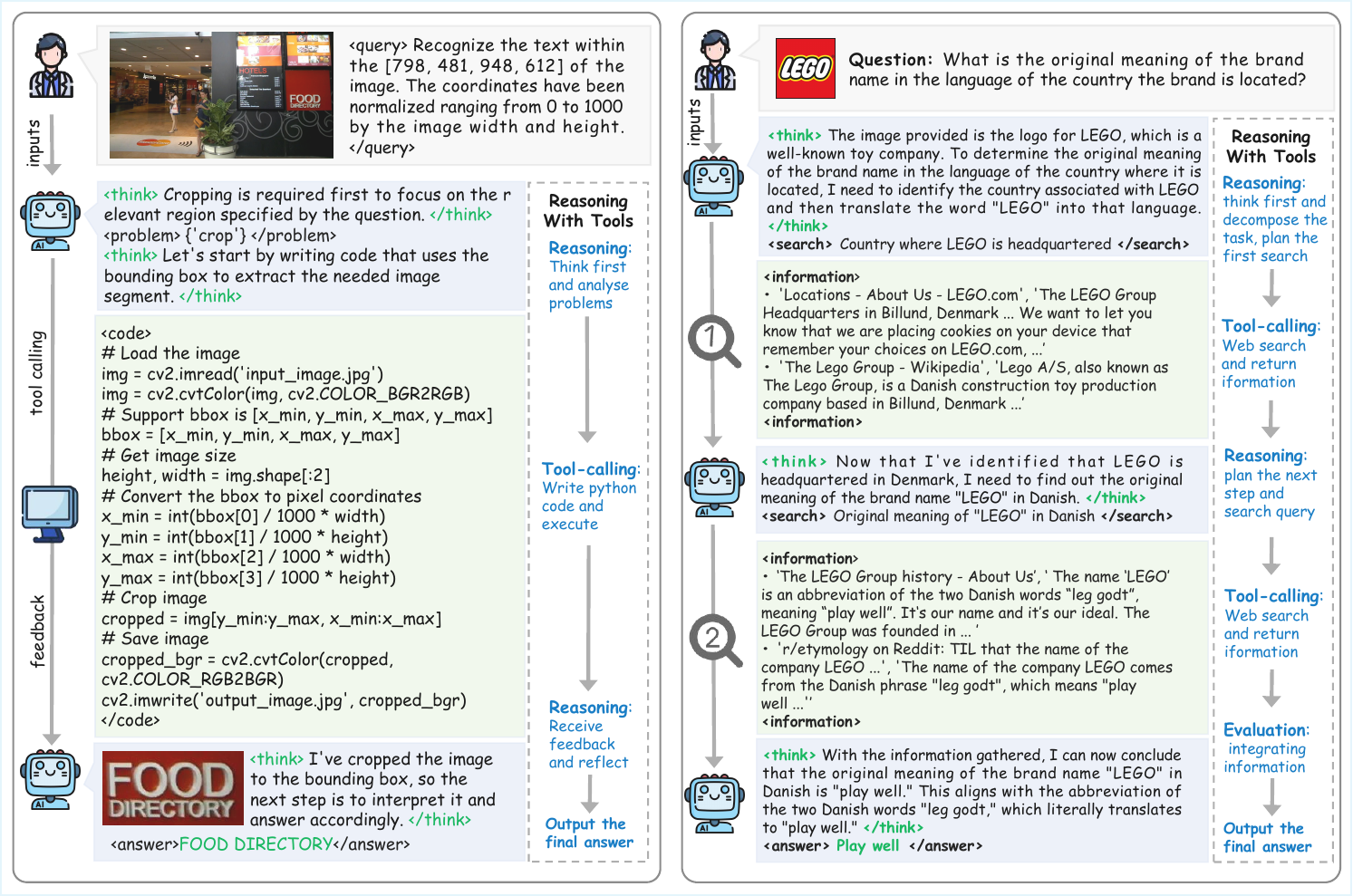}
   \vspace{-12pt}
    \caption{\small \textbf{Visualization Inference Cases of \methodname.} Demonstrating \methodname's multi-modal agentic capabilities: processing an image and answering a question via code generation and execution (\textbf{left}), and solving multi-hop VQA through query decomposition and search tool invocation (\textbf{right}).
}
    \label{fig:case_mat_infere}
    \vspace{-12pt}
\end{figure}

\noindent \textbf{Visualization Results.} We present several visualization cases of applying \methodname on Qwen2.5-VL in Fig.~\ref{fig:case_mat_infere}.
The left side demonstrates the model first generating and executing Python code to process the image and extract relevant information to answer the question.
The right side showcases how the model decomposes the complex query and uses the search tool to gather necessary information before formulating the final answer.
Both examples detail the intermediate ``think'' and ``tool calling,'' steps involved in arriving at the final answer. Please refer to the \cref{appendix:more_cases} for more visualization examples.

%% file: sec/5_conclusion.tex
\section{Conclusion} \label{sec_5_conclusion}
In this work, we propose Visual Agentic Reinforcement Fine-Tuning (\methodname), an effective framework for training multimodal agents through RFT with verifiable rewards. Our method enables LVLMs to perform complex reasoning and tool-use behaviors by interacting with external environments in both multimodal agentic search and agentic coding scenarios. To support training and evaluation, we introduce the Multimodal Agentic Tool Bench (\benchname), which provides high-quality benchmarks for assessing tool-driven reasoning in multimodal contexts. \methodname achieves substantial performance improvements on the \benchname, outperforming proprietary models like GPT-4o, and generalizes well to existing multi-hop QA tasks such as 2Wiki and HotpotQA. These results present a promising direction for developing open-source, o3-style multimodal AI agents with strong reasoning and tool-use capabilities.

%% file: sec/6_appendix.tex
\section{Prompt Used}\label{sec:appendix_prompt}

\subsection{Prompt for Tool-Use}
In both agentic search and agentic coding tasks, we provide explicit and carefully structured prompts to guide the Large Vision-Language Models (LVLMs) in using external tools to solve complex multimodal problems. These prompts are crucial to ensuring that the model understands not only what tools are available, but also how and when to use them, and in what format the results should be returned.

Each prompt is composed of several key components:

Tool Description Block: We describe each available tool (e.g., a search engine, a code execution environment) in natural language, including its purpose, usage method, and input/output interface. For example, in the search setting, we explain that the `<search>' tag is used to query external knowledge sources, and that results will be returned as plaintext snippets in `<information>' tag.

Instructional Template: We include explicit instructions that tell the model how to structure its reasoning process. We enforce a standardized format that separates the model’s internal reasoning (within <think>...</think>) from its tool invocations (e.g., <search>...</search> or <code>...</code>), and finally from its answer (<answer>...</answer>). This structure helps reinforce step-by-step planning and decision execution.

Formatting Hints: The prompt reminds the model to strictly follow the expected format, which is essential for downstream reward computation (e.g., verifiable reward for answer matching or search content evaluation). This includes avoiding unnecessary explanation in <answer>, keeping search queries concise, and writing complete and executable Python code.

\begin{figure}[t]
    \centering
   \includegraphics[width=1.\linewidth]{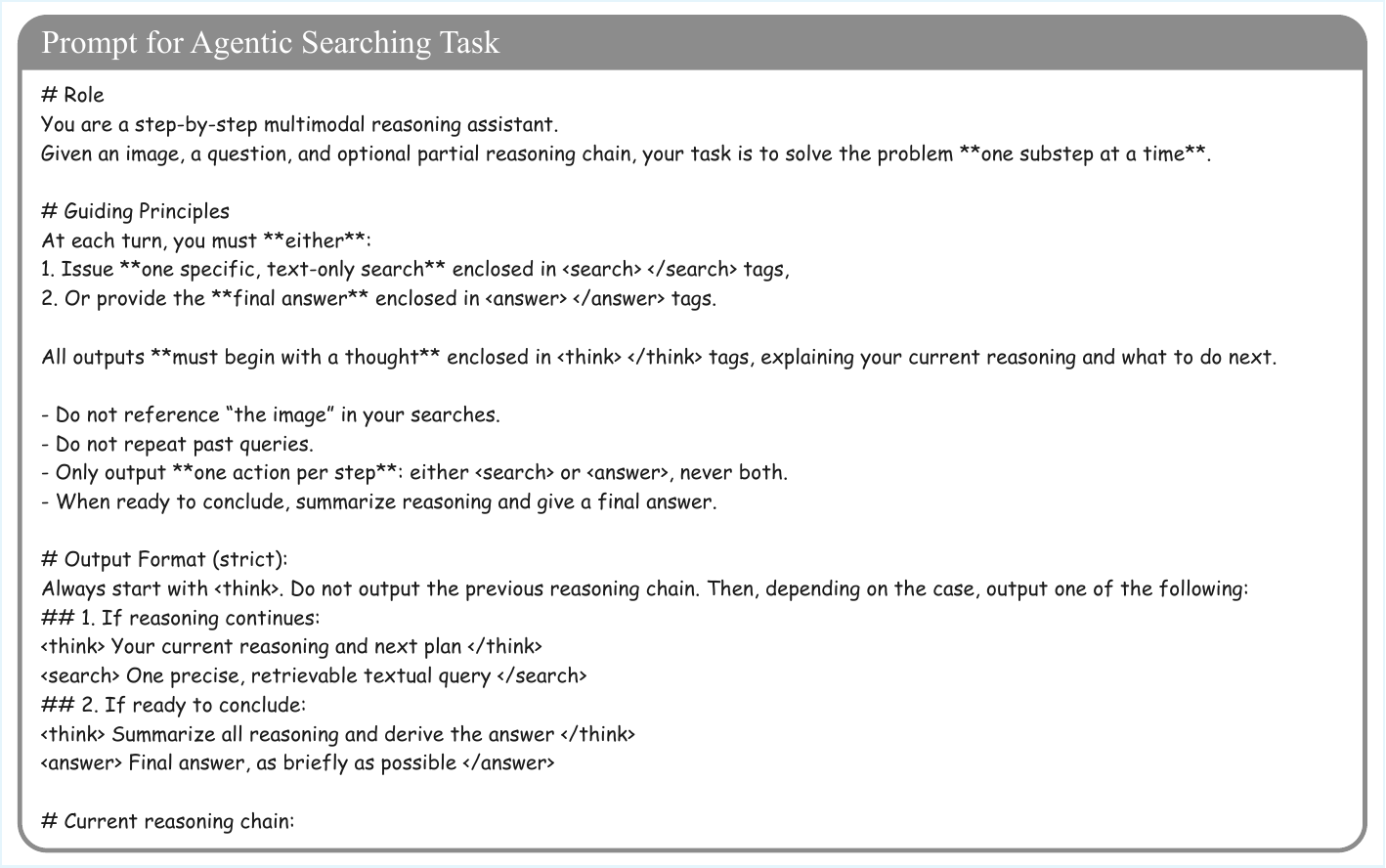}
    \caption{\small Prompt for Agentic Searching Tasks}
    \label{fig:prompt_for_search}
    \vspace{-8pt}
\end{figure}

\begin{figure}[t]
    \centering
   \includegraphics[width=1.\linewidth]{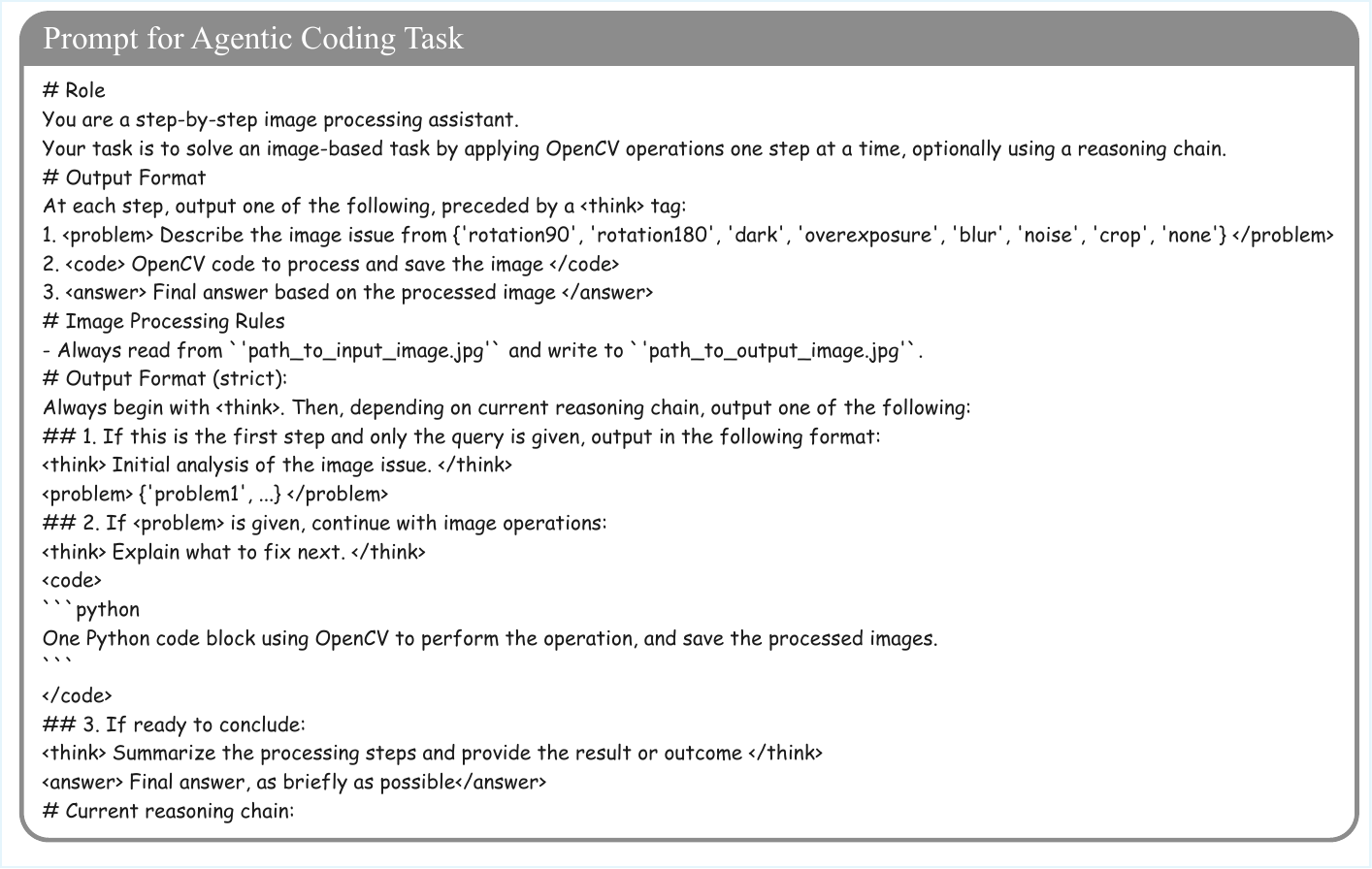}
    \caption{\small Prompt for Agentic Coding Tasks}
    \label{fig:prompt_for_coding}
    \vspace{-8pt}
\end{figure}

Complete examples of prompts used for each task are provided in Fig.\ref{fig:prompt_for_search} and Fig.\ref{fig:prompt_for_coding}. 

\subsection{Prompt for Baseline Evaluation}
We evaluate baseline models on \benchname-Bench using a single-turn QA setup, as they lack the ability to invoke tools. However, we find that some baseline models, such as XComposer2.5~\cite{xcomposer2.5}, tend to provide elaborative responses, occasionally including more explanation than necessary for direct-answer evaluation. These outputs make it difficult to extract clean answers, resulting in artificially low F1 and Exact Match (EM) scores during automatic evaluation.

To address this issue, we apply a lightweight instruction intervention: we append a direct-answer prompt suffix to the end of each test question, explicitly instructing the model to produce a concise answer. This strategy helps align the model’s output with the expected format used in our automatic scoring pipeline.

A commonly used suffix is: `Answer the question directly.'

This addition encourages models to avoid verbose explanation and return a short, focused response. While simple, this modification significantly improves evaluation compatibility and helps ensure a fairer comparison across models with varying instruction-following capabilities.

\begin{figure}[t]
    \centering
   \includegraphics[width=1.\linewidth]{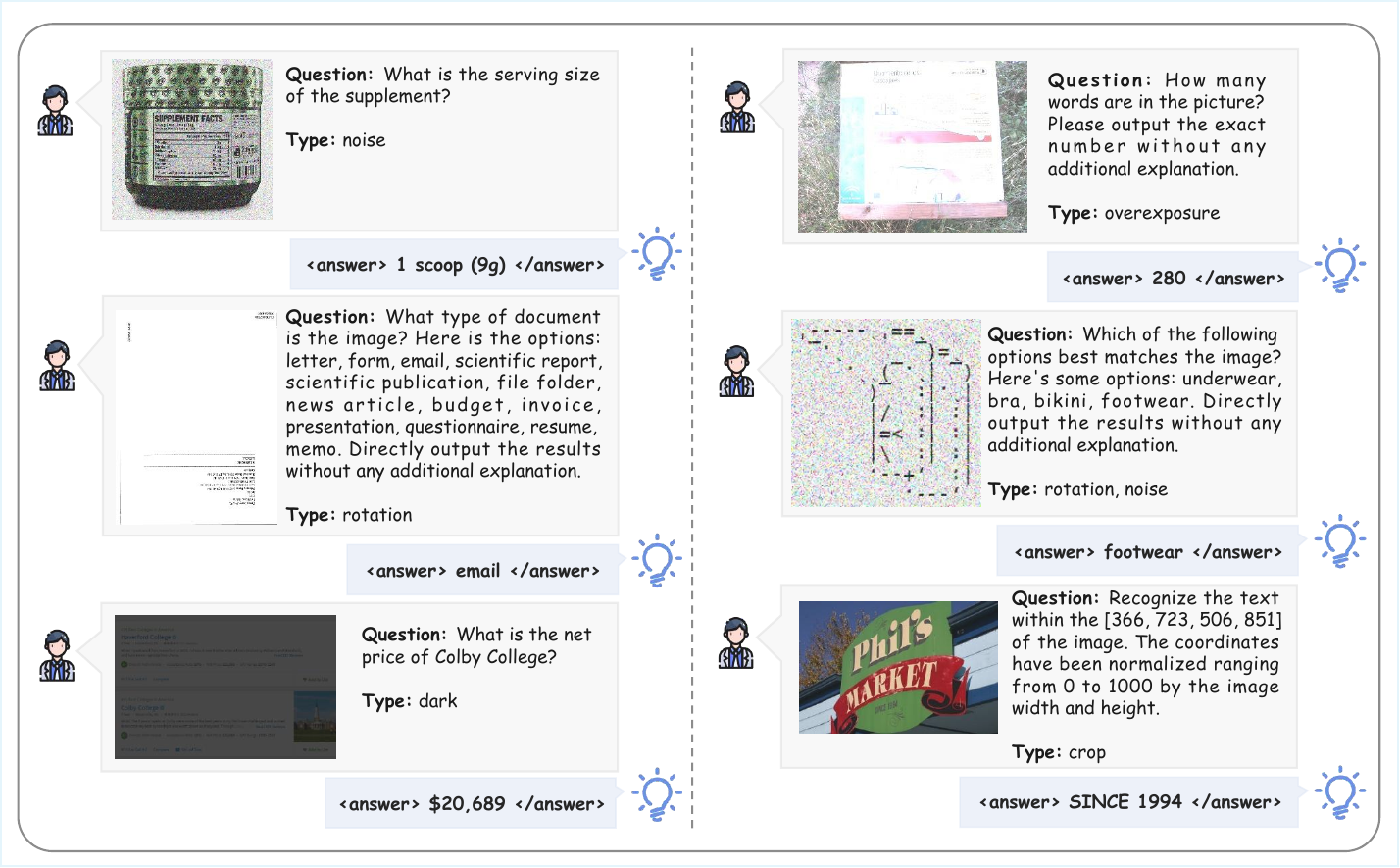}
    \caption{\small \codename Data Examples}
    \label{fig:mat_coding_case}
\end{figure}

\begin{figure}[t]
    \centering
   \includegraphics[width=1.\linewidth]{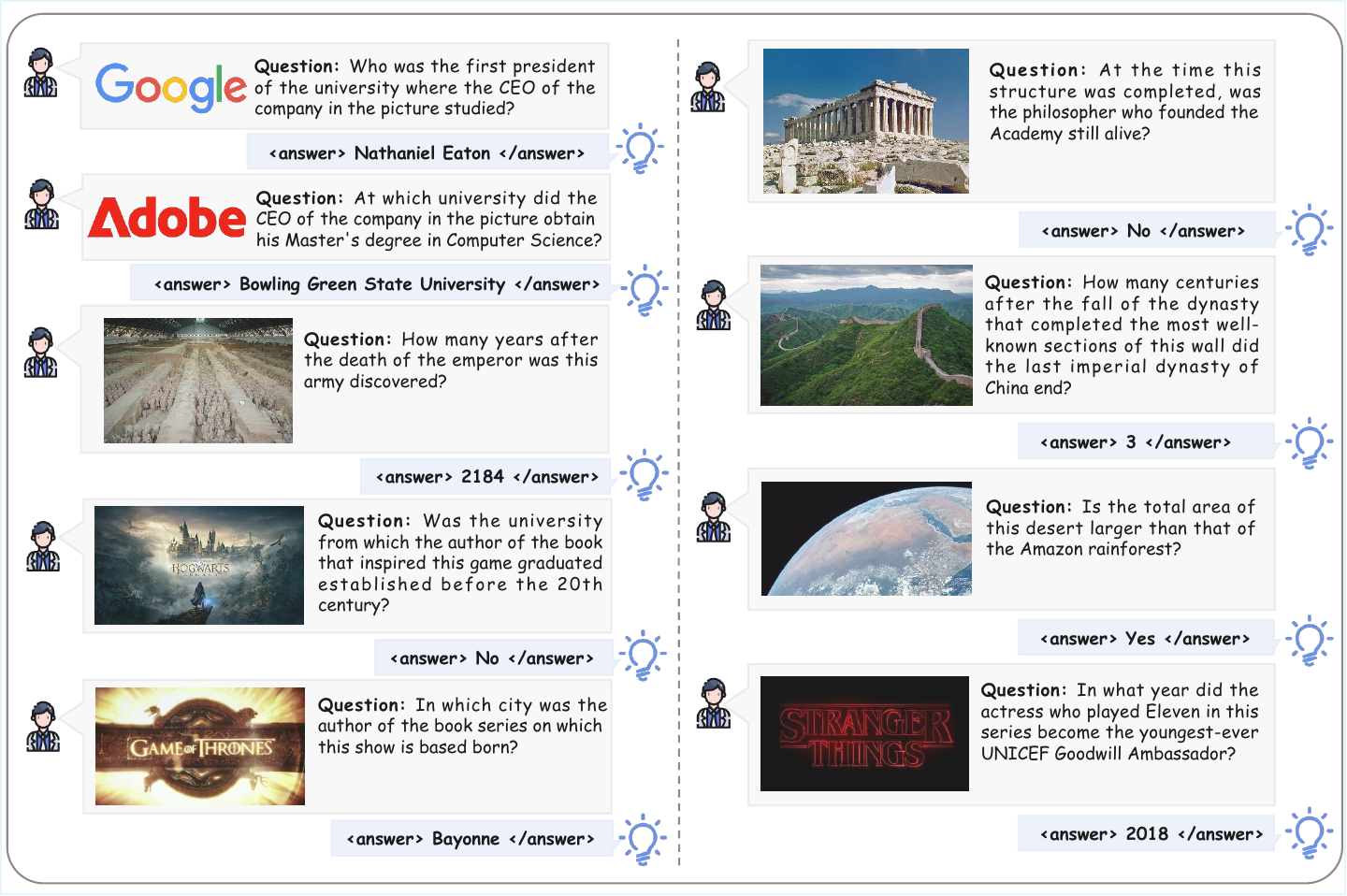}
    \caption{\small \searchname Data Examples}
    \label{fig:mat_search_case}
\end{figure}

\begin{figure}[t]
    \centering
   \includegraphics[width=1.\linewidth]{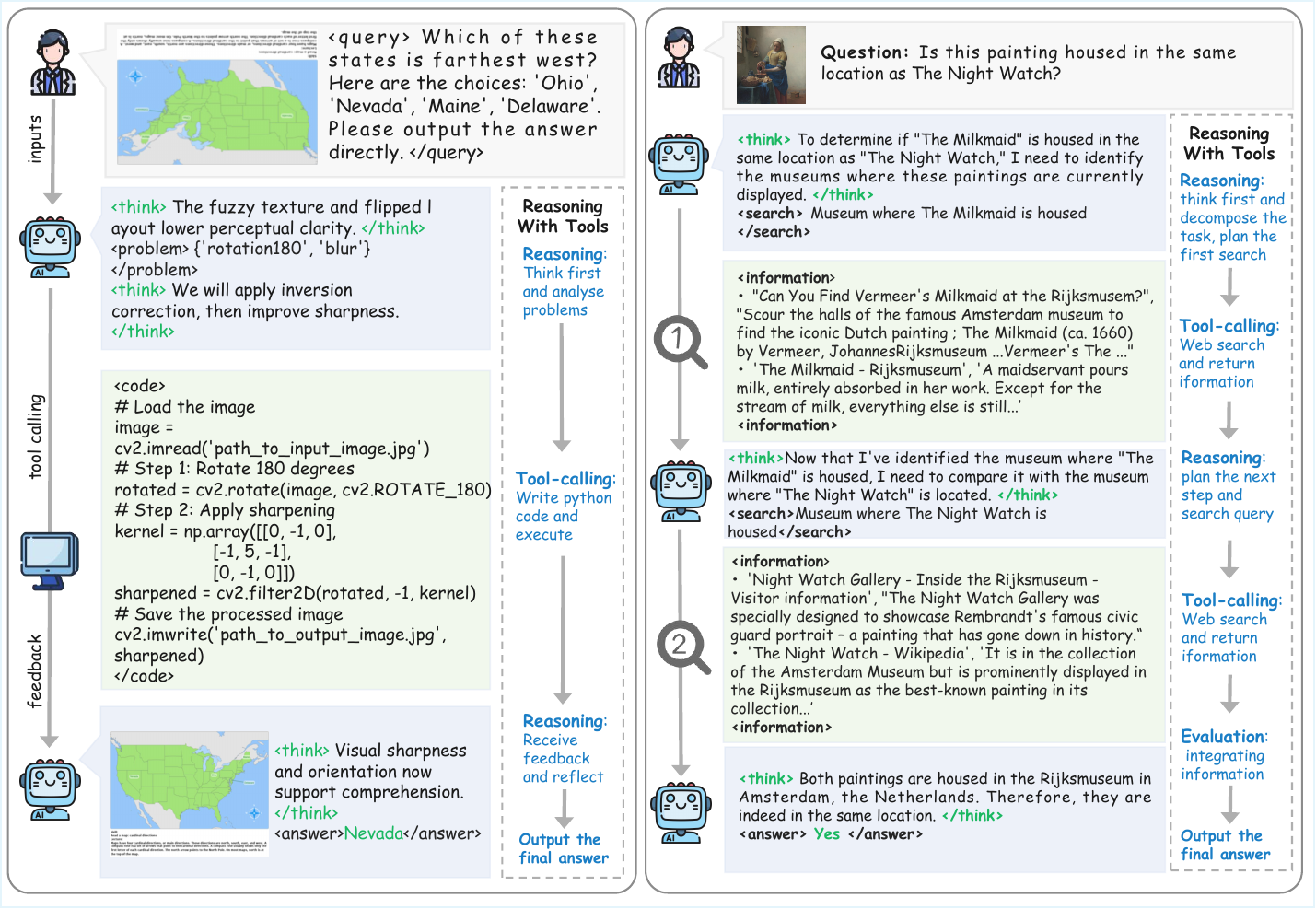}
    \caption{\small Examples of \methodname Inference on \benchname.}
    \label{fig:sup_mat_indere}
\end{figure}

\section{Training Data and Benchmark}\label{sec:appendix_datas}

\subsection{Data Source}
The Multimodal Agentic Tool Bench (\benchname) and the associated training datasets constructed in this work are derived from a combination of human-curated data and publicly available benchmarks. We design \benchname specifically to support the evaluation of agentic reasoning and tool-use behavior in large vision-language models (LVLMs), covering both search-based and code-based multimodal tasks.

For the Agentic Search task, we construct a high-quality dataset consisting of manually collected image-text pairs paired with hand-crafted multi-hop questions and their corresponding answers. The data is designed to require reasoning over both visual and textual information, and often necessitates decomposing the question and retrieving missing knowledge externally. This dataset forms the basis of both the training split and the \searchname benchmark, which we use to evaluate the model's ability to plan, decompose tasks, and invoke search tools.

For the Agentic Coding task, we curate examples from a wide range of established multimodal data source, including: OCRBench-v2~\cite{fu2024ocrbench}, RICO~\cite{deka2017rico}, MSRA-TD500~\cite{msra}, TextBookQA~\cite{TextBookQA}, OCRVQA~\cite{fu2024ocrbench}, MathVision~\cite{mathew2021docvqa}, DocVQA~\cite{mathew2021docvqa}, HierText~\cite{hiertext}, and FUNSD~\cite{jaume2019funsd}. 

We extract or design visual question-answering samples from these datasets and apply various distortions (e.g., rotation, blur, brightness variation) to the input images, thereby creating challenging scenarios that require the model to write and execute code to preprocess the image before answering the question. These examples serve both as training data for \methodname and as part of the \codename benchmark, allowing us to evaluate the model’s ability to reason about visual data, plan tool use, and manipulate inputs through code execution.

\subsection{Data Details}
In this section, we elaborate on the datasets constructed and used in our work, including overall data volume, data categories, task-specific splits. These datasets serve as the foundation for training and evaluating \methodname on two key scenarios: \textbf{agentic search} and \textbf{agentic coding}.

\subsubsection{Dataset Volume}
We construct both training and evaluation data for the two agentic tasks.

\textbf{For Agentic Search}, we manually annotate: \textit{20 training samples}, each accompanied by intermediate reasoning steps and structured tool calls, to teach the model how to decompose and retrieve relevant information. \textit{150 test samples} that comprise the \searchname benchmark. Each test case includes a question, an answer, and supporting reference material that may contain both relevant and distractor content. These test questions require multi-hop reasoning and often tool-assisted retrieval.

\textbf{For Agentic Coding}, we create a larger dataset due to the novelty and complexity of the task for base models: \textit{1,200 training examples} covering a wide range of image distortions and coding requirements. \textit{200 test examples} that form the \codename benchmark. These examples are manually validated to ensure the benchmark's quality. All test images in \codename have been manually reviewed to ensure high quality, balanced coverage across different types.

\subsubsection{Data Categorization}
\paragraph{\searchname.}
All examples in \searchname are constructed to require \textbf{multimodal, multi-hop reasoning}, where the question cannot be easily answered from the image or text alone. The questions typically involve factual knowledge, visual element, and indirect relationships, prompting the model to decompose the problem and retrieve intermediate facts via a search tool.

\paragraph{\codename.}
To simulate real-world scenarios where visual input quality varies, we construct coding-based VQA examples that include a mix of visually degraded images, clean (non-degraded) images, and images that require cropping-based preprocessing to locate relevant content. We synthetically apply the following types of distortions:

\textit{Rotation}: 90° or 180° clockwise.

\textit{Lighting}: including darkening and overexposure.

\textit{Blur}: Gaussian blur with varied kernel sizes.

\textit{Noise}: Noise with random standard deviation.

\textit{None}: clean images with no distortion.

\textit{Crop}: Requires removing irrelevant regions or focusing on a subregion of interest.

We further introduce \textbf{compositional distortions}, where several types (e.g., rotation + dark, blur + noise) are applied simultaneously, increasing the complexity of the required code logic.

\subsubsection{Distribution}
In \codename test set: Each single distortion category (e.g., \texttt{rotation90}, \texttt{blur}, etc.) contains \textit{10 examples}. \textit{Compositional distortions} and \textit{crop cases} make up the remaining \textit{130 examples}, ensuring the test set covers both simple and hard cases.

In the 1,200 training examples: Each category (including all single distortions, compositional combinations, and clean images) is evenly represented with \textit{100 examples} each.This ensures balanced exposure to diverse visual conditions and tool-use requirements.

\subsubsection{Difficulty Split: Simple vs. Hard}

To facilitate finer-grained analysis, we divide both \searchname and \codename benchmarks into Simple and Hard subsets based on the required reasoning complexity and tool invocation difficulty.

\paragraph{\searchname:} \textit{Simple} samples involve less reasoning steps and minimal ambiguity in evidence retrieval. \textit{Hard} samples require more reasoning hops, include distractor references, or involve indirect logical connections that challenge decomposition and retrieval abilities.

\paragraph{\codename:} \textit{Simple} samples include cases with a single distortion (e.g., only rotation or only blur), which are relatively easier to identify and correct via code. \textit{Hard} samples involve: (1) Multiple distortions applied simultaneously, (2) Complex visual layouts requiring cropping. This difficulty split is useful for evaluating the robustness of agentic reasoning and the model’s capacity to invoke the correct tools under varying levels of uncertainty and visual complexity.

\subsubsection{Dataset Cases} \label{appendix:more_cases}
We present several data examples from \codename in Fig.~\ref{fig:mat_coding_case}. Examples from \searchname are shown in Fig.~\ref{fig:mat_search_case}.
Additionally, we present several inference examples of \methodname from \benchname in Fig.~\ref{fig:sup_mat_indere}.

\section{Limitation} \label{sec:limitaion}
While our work demonstrates strong results, we acknowledge two limitations. (1) Our current implementation of \methodname focuses on two representative multimodal agentic tasks—search and coding—but does not cover other tool-augmented use cases. More tools will be supported in the future. (2) The \benchname benchmark, while diverse, is relatively small in scale compared to general-purpose datasets. We hope future work can expand both the range of agentic tools and the scale of evaluation benchmarks to further explore these directions.

\section{Potential Societal Impact} \label{app:impact}

This work explores reinforcement fine-tuning for multimodal agentic systems, enabling large vision-language models (LVLMs) to reason, decompose tasks, and interact with external tools in dynamic real-world settings. On the positive side, such capabilities could significantly enhance AI assistants in education, scientific analysis, and complex decision support, especially in domains requiring reasoning over visual content and structured tool use. For instance, improved agentic coding and search could support visually impaired users through adaptive visual understanding or assist professionals in automating complex workflows.

However, the deployment of tool-augmented agents also presents societal risks. If not properly constrained, such systems may misuse external tools (e.g., search engines or code execution) to produce unsafe outputs.

%% file: neurips_2025.bbl
\begin{thebibliography}{10}

\bibitem{bai2025qwen25vl}
Shuai Bai, Keqin Chen, Xuejing Liu, Jialin Wang, Wenbin Ge, Sibo Song, Kai Dang, Peng Wang, Shijie Wang, Jun Tang, et~al.
\newblock Qwen2. 5-vl technical report.
\newblock {\em arXiv preprint arXiv:2502.13923}, 2025.

\bibitem{researcher}
Mingyang Chen, Tianpeng Li, Haoze Sun, Yijie Zhou, Chenzheng Zhu, Haofen Wang, Jeff~Z Pan, Wen Zhang, Huajun Chen, Fan Yang, et~al.
\newblock Learning to reason with search for llms via reinforcement learning.
\newblock {\em arXiv preprint arXiv:2503.19470}, 2025.

\bibitem{internvl2.5}
Zhe Chen, Weiyun Wang, Yue Cao, Yangzhou Liu, Zhangwei Gao, Erfei Cui, Jinguo Zhu, Shenglong Ye, Hao Tian, Zhaoyang Liu, et~al.
\newblock Expanding performance boundaries of open-source multimodal models with model, data, and test-time scaling.
\newblock {\em arXiv preprint arXiv:2412.05271}, 2024.

\bibitem{deka2017rico}
Biplab Deka, Zifeng Huang, Chad Franzen, Joshua Hibschman, Daniel Afergan, Yang Li, Jeffrey Nichols, and Ranjitha Kumar.
\newblock Rico: A mobile app dataset for building data-driven design applications.
\newblock In {\em Proceedings of the 30th annual ACM symposium on user interface software and technology}, pages 845--854, 2017.

\bibitem{feng2025retool}
Jiazhan Feng, Shijue Huang, Xingwei Qu, Ge~Zhang, Yujia Qin, Baoquan Zhong, Chengquan Jiang, Jinxin Chi, and Wanjun Zhong.
\newblock Retool: Reinforcement learning for strategic tool use in llms.
\newblock {\em arXiv preprint arXiv:2504.11536}, 2025.

\bibitem{fu2024ocrbench}
Ling Fu, Biao Yang, Zhebin Kuang, Jiajun Song, Yuzhe Li, Linghao Zhu, Qidi Luo, Xinyu Wang, Hao Lu, Mingxin Huang, et~al.
\newblock Ocrbench v2: An improved benchmark for evaluating large multimodal models on visual text localization and reasoning.
\newblock {\em arXiv preprint arXiv:2501.00321}, 2024.

\bibitem{goldie2025synthetic}
Anna Goldie, Azalia Mirhoseini, Hao Zhou, Irene Cai, and Christopher~D Manning.
\newblock Synthetic data generation \& multi-step rl for reasoning \& tool use.
\newblock {\em arXiv preprint arXiv:2504.04736}, 2025.

\bibitem{guo2025deepseek-r1}
Daya Guo, Dejian Yang, Haowei Zhang, Junxiao Song, Ruoyu Zhang, Runxin Xu, Qihao Zhu, Shirong Ma, Peiyi Wang, Xiao Bi, et~al.
\newblock Deepseek-r1: Incentivizing reasoning capability in llms via reinforcement learning.
\newblock {\em arXiv preprint arXiv:2501.12948}, 2025.

\bibitem{ho2020constructing2wiki}
Xanh Ho, Anh-Khoa~Duong Nguyen, Saku Sugawara, and Akiko Aizawa.
\newblock Constructing a multi-hop qa dataset for comprehensive evaluation of reasoning steps.
\newblock {\em arXiv preprint arXiv:2011.01060}, 2020.

\bibitem{hurst2024gpt4o}
Aaron Hurst, Adam Lerer, Adam~P Goucher, Adam Perelman, Aditya Ramesh, Aidan Clark, AJ~Ostrow, Akila Welihinda, Alan Hayes, Alec Radford, et~al.
\newblock Gpt-4o system card.
\newblock {\em arXiv preprint arXiv:2410.21276}, 2024.

\bibitem{jaech2024openaio1}
Aaron Jaech, Adam Kalai, Adam Lerer, Adam Richardson, Ahmed El-Kishky, Aiden Low, Alec Helyar, Aleksander Madry, Alex Beutel, Alex Carney, et~al.
\newblock Openai o1 system card.
\newblock {\em arXiv preprint arXiv:2412.16720}, 2024.

\bibitem{jaume2019funsd}
Guillaume Jaume, Hazim~Kemal Ekenel, and Jean-Philippe Thiran.
\newblock Funsd: A dataset for form understanding in noisy scanned documents.
\newblock In {\em 2019 International Conference on Document Analysis and Recognition Workshops (ICDARW)}, volume~2, pages 1--6. IEEE, 2019.

\bibitem{jin2025search-r1}
Bowen Jin, Hansi Zeng, Zhenrui Yue, Jinsung Yoon, Sercan Arik, Dong Wang, Hamed Zamani, and Jiawei Han.
\newblock Search-r1: Training llms to reason and leverage search engines with reinforcement learning.
\newblock {\em arXiv preprint arXiv:2503.09516}, 2025.

\bibitem{TextBookQA}
Aniruddha Kembhavi, Minjoon Seo, Dustin Schwenk, Jonghyun Choi, Ali Farhadi, and Hannaneh Hajishirzi.
\newblock Are you smarter than a sixth grader? textbook question answering for multimodal machine comprehension.
\newblock In {\em Proceedings of the IEEE Conference on Computer Vision and Pattern recognition}, pages 4999--5007, 2017.

\bibitem{lambert2024ttulu}
Nathan Lambert, Jacob Morrison, Valentina Pyatkin, Shengyi Huang, Hamish Ivison, Faeze Brahman, Lester James~V Miranda, Alisa Liu, Nouha Dziri, Shane Lyu, et~al.
\newblock T$\backslash$" ulu 3: Pushing frontiers in open language model post-training.
\newblock {\em arXiv preprint arXiv:2411.15124}, 2024.

\bibitem{li2024llavaov}
Bo~Li, Yuanhan Zhang, Dong Guo, Renrui Zhang, Feng Li, Hao Zhang, Kaichen Zhang, Peiyuan Zhang, Yanwei Li, Ziwei Liu, et~al.
\newblock Llava-onevision: Easy visual task transfer.
\newblock {\em arXiv preprint arXiv:2408.03326}, 2024.

\bibitem{li2024llavanext}
Feng Li, Renrui Zhang, Hao Zhang, Yuanhan Zhang, Bo~Li, Wei Li, Zejun Ma, and Chunyuan Li.
\newblock Llava-next-interleave: Tackling multi-image, video, and 3d in large multimodal models.
\newblock {\em arXiv preprint arXiv:2407.07895}, 2024.

\bibitem{li2025search-o1}
Xiaoxi Li, Guanting Dong, Jiajie Jin, Yuyao Zhang, Yujia Zhou, Yutao Zhu, Peitian Zhang, and Zhicheng Dou.
\newblock Search-o1: Agentic search-enhanced large reasoning models.
\newblock {\em arXiv preprint arXiv:2501.05366}, 2025.

\bibitem{li2025webthinker}
Xiaoxi Li, Jiajie Jin, Guanting Dong, Hongjin Qian, Yutao Zhu, Yongkang Wu, Ji-Rong Wen, and Zhicheng Dou.
\newblock Webthinker: Empowering large reasoning models with deep research capability.
\newblock {\em arXiv preprint arXiv:2504.21776}, 2025.

\bibitem{li2025torl}
Xuefeng Li, Haoyang Zou, and Pengfei Liu.
\newblock Torl: Scaling tool-integrated rl.
\newblock {\em arXiv preprint arXiv:2503.23383}, 2025.

\bibitem{liao2024mario}
Minpeng Liao, Wei Luo, Chengxi Li, Jing Wu, and Kai Fan.
\newblock Mario: Math reasoning with code interpreter output--a reproducible pipeline.
\newblock {\em arXiv preprint arXiv:2401.08190}, 2024.

\bibitem{liu2024deepseekv3}
Aixin Liu, Bei Feng, Bing Xue, Bingxuan Wang, Bochao Wu, Chengda Lu, Chenggang Zhao, Chengqi Deng, Chenyu Zhang, Chong Ruan, et~al.
\newblock Deepseek-v3 technical report.
\newblock {\em arXiv preprint arXiv:2412.19437}, 2024.

\bibitem{liu2024skywork}
Chris~Yuhao Liu, Liang Zeng, Jiacai Liu, Rui Yan, Jujie He, Chaojie Wang, Shuicheng Yan, Yang Liu, and Yahui Zhou.
\newblock Skywork-reward: Bag of tricks for reward modeling in llms.
\newblock {\em arXiv preprint arXiv:2410.18451}, 2024.

\bibitem{llava1.5}
Haotian Liu, Chunyuan Li, Yuheng Li, and Yong~Jae Lee.
\newblock Improved baselines with visual instruction tuning.
\newblock In {\em Proceedings of the IEEE/CVF Conference on Computer Vision and Pattern Recognition}, pages 26296--26306, 2024.

\bibitem{liu2025visualrft}
Ziyu Liu, Zeyi Sun, Yuhang Zang, Xiaoyi Dong, Yuhang Cao, Haodong Duan, Dahua Lin, and Jiaqi Wang.
\newblock Visual-rft: Visual reinforcement fine-tuning.
\newblock {\em arXiv preprint arXiv:2503.01785}, 2025.

\bibitem{liu2024mia}
Ziyu Liu, Yuhang Zang, Xiaoyi Dong, Pan Zhang, Yuhang Cao, Haodong Duan, Conghui He, Yuanjun Xiong, Dahua Lin, and Jiaqi Wang.
\newblock Mia-dpo: Multi-image augmented direct preference optimization for large vision-language models.
\newblock {\em arXiv preprint arXiv:2410.17637}, 2024.

\bibitem{hiertext}
Shangbang Long, Siyang Qin, Dmitry Panteleev, Alessandro Bissacco, Yasuhisa Fujii, and Michalis Raptis.
\newblock Towards end-to-end unified scene text detection and layout analysis.
\newblock In {\em Proceedings of the IEEE/CVF Conference on Computer Vision and Pattern Recognition}, pages 1049--1059, 2022.

\bibitem{mathew2021docvqa}
Minesh Mathew, Dimosthenis Karatzas, and CV~Jawahar.
\newblock Docvqa: A dataset for vqa on document images.
\newblock In {\em Proceedings of the IEEE/CVF winter conference on applications of computer vision}, pages 2200--2209, 2021.

\bibitem{OpenAI-o3}
OpenAI.
\newblock Openai o3 and o4-mini system card.
\newblock \url{https://openai.com/index/o3-o4-mini-system-card/}, 2025.
\newblock Accessed: 2025-04-16.

\bibitem{ouyang2022training}
Long Ouyang, Jeffrey Wu, Xu~Jiang, Diogo Almeida, Carroll Wainwright, Pamela Mishkin, Chong Zhang, Sandhini Agarwal, Katarina Slama, Alex Ray, et~al.
\newblock Training language models to follow instructions with human feedback.
\newblock {\em Advances in neural information processing systems}, 35:27730--27744, 2022.

\bibitem{bfcl3}
Shishir~G Patil, Tianjun Zhang, Xin Wang, and Joseph~E Gonzalez.
\newblock Gorilla: Large language model connected with massive apis.
\newblock {\em Advances in Neural Information Processing Systems}, 37:126544--126565, 2024.

\bibitem{press2022measuringbamboodle}
Ofir Press, Muru Zhang, Sewon Min, Ludwig Schmidt, Noah~A Smith, and Mike Lewis.
\newblock Measuring and narrowing the compositionality gap in language models.
\newblock {\em arXiv preprint arXiv:2210.03350}, 2022.

\bibitem{qian2025toolrl}
Cheng Qian, Emre~Can Acikgoz, Qi~He, Hongru Wang, Xiusi Chen, Dilek Hakkani-T{\"u}r, Gokhan Tur, and Heng Ji.
\newblock Toolrl: Reward is all tool learning needs.
\newblock {\em arXiv preprint arXiv:2504.13958}, 2025.

\bibitem{dpo}
Rafael Rafailov, Archit Sharma, Eric Mitchell, Christopher~D Manning, Stefano Ermon, and Chelsea Finn.
\newblock Direct preference optimization: Your language model is secretly a reward model.
\newblock {\em Advances in Neural Information Processing Systems}, 36:53728--53741, 2023.

\bibitem{ppo}
John Schulman, Filip Wolski, Prafulla Dhariwal, Alec Radford, and Oleg Klimov.
\newblock Proximal policy optimization algorithms.
\newblock {\em arXiv preprint arXiv:1707.06347}, 2017.

\bibitem{grpo}
Zhihong Shao, Peiyi Wang, Qihao Zhu, Runxin Xu, Junxiao Song, Xiao Bi, Haowei Zhang, Mingchuan Zhang, YK~Li, Y~Wu, et~al.
\newblock Deepseekmath: Pushing the limits of mathematical reasoning in open language models.
\newblock {\em arXiv preprint arXiv:2402.03300}, 2024.

\bibitem{singh2025agentic-RL-math}
Joykirat Singh, Raghav Magazine, Yash Pandya, and Akshay Nambi.
\newblock Agentic reasoning and tool integration for llms via reinforcement learning.
\newblock {\em arXiv preprint arXiv:2505.01441}, 2025.

\bibitem{song2025r1-searcher}
Huatong Song, Jinhao Jiang, Yingqian Min, Jie Chen, Zhipeng Chen, Wayne~Xin Zhao, Lei Fang, and Ji-Rong Wen.
\newblock R1-searcher: Incentivizing the search capability in llms via reinforcement learning.
\newblock {\em arXiv preprint arXiv:2503.05592}, 2025.

\bibitem{sun2025zerosearch}
Hao Sun, Zile Qiao, Jiayan Guo, Xuanbo Fan, Yingyan Hou, Yong Jiang, Pengjun Xie, Fei Huang, and Yan Zhang.
\newblock Zerosearch: Incentivize the search capability of llms without searching.
\newblock {\em arXiv preprint arXiv:2505.04588}, 2025.

\bibitem{sun2023aligning}
Zhiqing Sun, Sheng Shen, Shengcao Cao, Haotian Liu, Chunyuan Li, Yikang Shen, Chuang Gan, Liang-Yan Gui, Yu-Xiong Wang, Yiming Yang, et~al.
\newblock Aligning large multimodal models with factually augmented rlhf.
\newblock {\em arXiv preprint arXiv:2309.14525}, 2023.

\bibitem{llavarlhf}
Zhiqing Sun, Sheng Shen, Shengcao Cao, Haotian Liu, Chunyuan Li, Yikang Shen, Chuang Gan, Liang-Yan Gui, Yu-Xiong Wang, Yiming Yang, et~al.
\newblock Aligning large multimodal models with factually augmented rlhf.
\newblock {\em arXiv preprint arXiv:2309.14525}, 2023.

\bibitem{team2023gemini}
Gemini Team, Rohan Anil, Sebastian Borgeaud, Jean-Baptiste Alayrac, Jiahui Yu, Radu Soricut, Johan Schalkwyk, Andrew~M Dai, Anja Hauth, Katie Millican, et~al.
\newblock Gemini: a family of highly capable multimodal models.
\newblock {\em arXiv preprint arXiv:2312.11805}, 2023.

\bibitem{team2025kimi}
Kimi Team, Angang Du, Bofei Gao, Bowei Xing, Changjiu Jiang, Cheng Chen, Cheng Li, Chenjun Xiao, Chenzhuang Du, Chonghua Liao, et~al.
\newblock Kimi k1. 5: Scaling reinforcement learning with llms.
\newblock {\em arXiv preprint arXiv:2501.12599}, 2025.

\bibitem{trivedi2022musique}
Harsh Trivedi, Niranjan Balasubramanian, Tushar Khot, and Ashish Sabharwal.
\newblock Musique: Multihop questions via single-hop question composition.
\newblock {\em Transactions of the Association for Computational Linguistics}, 10:539--554, 2022.

\bibitem{wang2024qwen2vl}
Peng Wang, Shuai Bai, Sinan Tan, Shijie Wang, Zhihao Fan, Jinze Bai, Keqin Chen, Xuejing Liu, Jialin Wang, Wenbin Ge, et~al.
\newblock Qwen2-vl: Enhancing vision-language model's perception of the world at any resolution.
\newblock {\em arXiv preprint arXiv:2409.12191}, 2024.

\bibitem{yang2018hotpotqa}
Zhilin Yang, Peng Qi, Saizheng Zhang, Yoshua Bengio, William~W Cohen, Ruslan Salakhutdinov, and Christopher~D Manning.
\newblock Hotpotqa: A dataset for diverse, explainable multi-hop question answering.
\newblock {\em arXiv preprint arXiv:1809.09600}, 2018.

\bibitem{msra}
Cong Yao, Xiang Bai, Wenyu Liu, Yi~Ma, and Zhuowen Tu.
\newblock Detecting texts of arbitrary orientations in natural images.
\newblock In {\em 2012 IEEE conference on computer vision and pattern recognition}, pages 1083--1090. IEEE, 2012.

\bibitem{yu2024rlhfv}
Tianyu Yu, Yuan Yao, Haoye Zhang, Taiwen He, Yifeng Han, Ganqu Cui, Jinyi Hu, Zhiyuan Liu, Hai-Tao Zheng, Maosong Sun, et~al.
\newblock {RlHF-V}: Towards trustworthy mllms via behavior alignment from fine-grained correctional human feedback.
\newblock In {\em CVPR}, 2024.

\bibitem{yu2024rlaif}
Tianyu Yu, Haoye Zhang, Yuan Yao, Yunkai Dang, Da~Chen, Xiaoman Lu, Ganqu Cui, Taiwen He, Zhiyuan Liu, Tat-Seng Chua, et~al.
\newblock {RLAIF-V}: Aligning mllms through open-source ai feedback for super gpt-4v trustworthiness.
\newblock {\em arXiv preprint arXiv:2405.17220}, 2024.

\bibitem{zang2025internlm}
Yuhang Zang, Xiaoyi Dong, Pan Zhang, Yuhang Cao, Ziyu Liu, Shengyuan Ding, Shenxi Wu, Yubo Ma, Haodong Duan, Wenwei Zhang, et~al.
\newblock Internlm-xcomposer2. 5-reward: A simple yet effective multi-modal reward model.
\newblock {\em arXiv preprint arXiv:2501.12368}, 2025.

\bibitem{xcomposer2.5}
Pan Zhang, Xiaoyi Dong, Yuhang Zang, Yuhang Cao, Rui Qian, Lin Chen, Qipeng Guo, Haodong Duan, Bin Wang, Linke Ouyang, et~al.
\newblock Internlm-xcomposer-2.5: A versatile large vision language model supporting long-contextual input and output.
\newblock {\em arXiv preprint arXiv:2407.03320}, 2024.

\bibitem{zhang2025nemotron}
Shaokun Zhang, Yi~Dong, Jieyu Zhang, Jan Kautz, Bryan Catanzaro, Andrew Tao, Qingyun Wu, Zhiding Yu, and Guilin Liu.
\newblock Nemotron-research-tool-n1: Tool-using language models with reinforced reasoning.
\newblock {\em arXiv preprint arXiv:2505.00024}, 2025.

\bibitem{zhou2024aligning}
Yiyang Zhou, Chenhang Cui, Rafael Rafailov, Chelsea Finn, and Huaxiu Yao.
\newblock Aligning modalities in vision large language models via preference fine-tuning.
\newblock {\em arXiv preprint arXiv:2402.11411}, 2024.

\end{thebibliography}
